\documentclass[10pt,twocolumn,letterpaper]{article}
\usepackage[pagenumbers]{cvpr}
\usepackage{microtype}
\usepackage{mathtools}
\usepackage{multirow}
\usepackage{array}
\usepackage{float}
\usepackage{siunitx}
\makeatletter
\let\cvprOriginalLenToUnit\LenToUnit
\renewcommand{\LenToUnit}[1]{\expandafter\cvprOriginalLenToUnit\expandafter{\the\dimexpr#1\relax}}
\newcommand{\clearfullpage}{%
  \clearpage
  \global\@colroom\@colht
  \global\vsize\@colht
}
\makeatother
\newcommand{\R}{\mathbb{R}}
\newcommand{\T}{^{\mathsf T}}

\newcommand{\statee}{State--Euler}
\newcommand{\stater}{State--RK4}

\newcommand{\dd}{\mathrm{d}}
\newcommand{\Nm}{\mathrm{N\,m}}

\newcommand{\keytitle}{Beyond Geometric Closure: Measuring and Diagnosing Mechanical Consistency in Visual Dynamics}
\definecolor{cvprblue}{rgb}{0.21,0.49,0.74}
\usepackage[pagebackref,breaklinks,colorlinks,allcolors=cvprblue]{hyperref}
\usepackage[capitalize,nameinlink]{cleveref}
\hypersetup{pdftitle={Beyond Geometric Closure: Measuring and Diagnosing Mechanical Consistency in Visual Dynamics},
            pdfauthor={Qibin Gu, Yunfeng Yan},pdfsubject={Independent physical evaluation of visual dynamics}}

\usepackage{placeins}
\title{\keytitle}
\author{Qibin Gu\\
Nanjing Tech University\\
{\small\href{mailto:guqibin123@njtech.edu.cn}{\texttt{guqibin123@njtech.edu.cn}}}
\and
Yunfeng Yan\\
Zhejiang University\\
{\small\href{mailto:yyff@zju.edu.cn}{\texttt{yyff@zju.edu.cn}}}}
\makeatletter
\patchcmd{\@maketitle}{\vspace*{12pt}}{%
  {\small\url{https://github.com/BIGPAO123321/mechanical-consistency-visual-dynamics}\par}%
  \vspace*{6pt}%
}{}{\PackageError{arxiv}{Could not add repository link to title block}{}}
\makeatother
\begin{document}
\setlength{\dbltextfloatsep}{10pt plus 2pt minus 2pt}
\setlength{\textfloatsep}{10pt plus 2pt minus 2pt}
\maketitle

\begin{abstract}
Visual accuracy, geometric validity, and agreement with applied forces are distinct requirements for visual dynamics. We introduce a diagnostic framework combining reaction-free SI residuals, reference-replay calibration, and controlled interventions on coordinate readout, training integration, and actuator response. A 343-assembly planar corpus provides 36 held-out test identities and 324 forecasts per setting across three seeds. We evaluate KeyCLD and CON at $64/256$ pixels, and V-JEPA + AC and LingBot-VA on 100 spatial test assemblies with three seeds. CON reduces image MSE by $37.6\%/37.2\%$ relative to persistence, while neither planar video model passes native geometry. Exact-state KeyCLD retains 100\% kinematic pass rates under Euler and RK4 training; point, torque, and energy errors fall by 84--88\%. Its continuous score rises from 52.65 to 64.30, with tracking and torque contributing 7.81 and 3.82 of the 11.65-point gain. Spatial CAD reconstruction enables force and energy measurements on 424--450 of 450 adapted forecasts per family and pathway; mean actuator residuals remain 5.4--16.6 times their targets. Matched reference replay and fixed-state probes separate coordinate recovery, numerical integration, and force-response calibration, establishing complementary diagnostics beyond visual and geometric accuracy.
\end{abstract}

\section{Introduction}
\label{sec:intro}
Mechanical consistency in visual prediction requires agreement with independently specified geometry, actuator inputs, and work in SI units. In rigid closed chains, closure fixes admissible shapes, whereas speed and acceleration along that path determine actuator demands. Exact-state KeyCLD illustrates this distinction: Euler and RK4 training both pass every geometric check, yet mean torque residuals fall from 4.56 to 0.72 N\,m (Table~\ref{tab:corpus}).

\begin{figure*}[t]
\centering
\includegraphics[width=\textwidth]{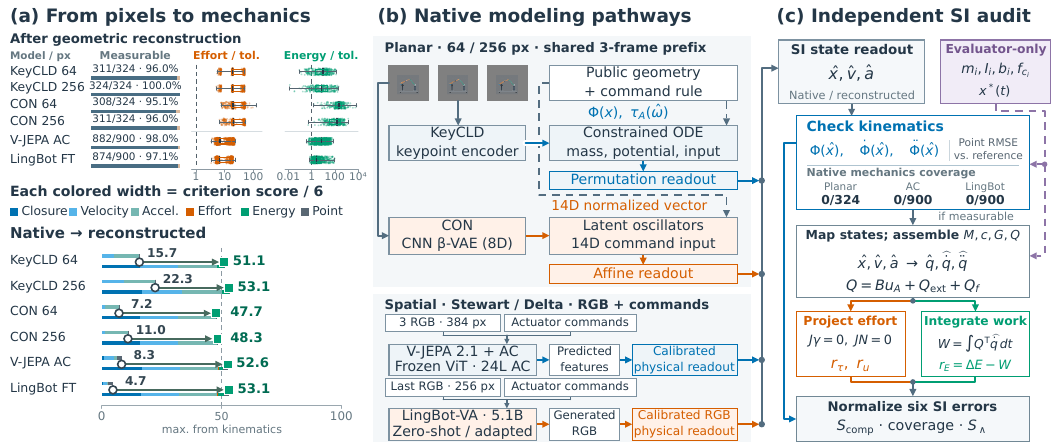}
\caption{Four pathways from pixels to mechanics. (a) After reconstruction, blue/tan bars show measurable/unavailable fractions, with measurable/total counts and percentages. Effort/energy plots show cases, quartiles, ranges, and dashed tolerances. Below, six contributions sum to $S_{\rm comp}$; arrows show reconstruction gains and 50 marks the maximum kinematic contribution. Scores use task-specific targets with balanced seeds, families, and conditions; each spatial pathway contains 900 cases (LingBot FT: 50 updates). (b,c) Physical readouts feed kinematic and effort/energy tests using independent SI parameters ($B u_A=e_1\tau_A$ for planar chains).}
\label{fig:audit}
\end{figure*}

Our framework links these residuals to their sources by combining reference-replay calibration with controlled interventions (\cref{fig:audit}). Readout changes, paired training integrators, and fixed-state torque perturbations distinguish coordinate, numerical, and constitutive errors. We apply the audit to constrained Lagrangian dynamics (KeyCLD)~\cite{daems2024keycld} and latent coupled oscillators (CON)~\cite{stolzle2024con} at 64 and 256 pixels, then to predictive features (V-JEPA 2.1 + AC)~\cite{murlabadia2026vjepa2_1,assran2025vjepa2} and generative video--action modeling (LingBot-VA)~\cite{li2026causal} on Stewart and Delta mechanisms.

We contribute (1) six SI criteria separating admissibility, actuator and energy balance, and tracking; (2) reference-calibrated readout, integrator, and force interventions that locate distinct error sources; and (3) planar and spatial benchmarks demonstrating these distinctions across four modeling paradigms.

\section{Related Work}
\label{sec:related}
\paragraph{Mechanical foundations.}
Hamiltonian~\cite{greydanus2019hamiltonian} and Lagrangian networks~\cite{cranmer2020lagrangian,lutter2019delan}, Cartesian constraints~\cite{finzi2020constraints}, and dissipative SymODEN~\cite{zhong2020dissipative} encode energy, geometry, and velocity-dependent forces. These priors and their analyses~\cite{gruver2022deconstructing} inform our diagnostics.

\paragraph{Learning physical evolution from images.}
Physion studies physical prediction from vision~\cite{bear2021physion}; Hamiltonian Generative Networks~\cite{toth2020hgn} and coordinate-aware Lagrangian models~\cite{zhong2020images} link images to mechanics. We examine constrained keypoints, latent oscillators, predictive features, and video--action generation.

\paragraph{Visual physics evaluation.}
Physics-IQ measures continuation fidelity~\cite{motamed2026physicsiq}, while MechVerse tests coupling between mechanical parts~\cite{jain2026mechverse}. Morpheus evaluates governing equations and physical invariants~\cite{tragoudaras2026morpheus}; WorldBench estimates calibrated parameters~\cite{upadhyay2026worldbench}, and Last Exam tests scale-free physical relationships~\cite{gao2026lastexam}. GAUGE's video-model branch separates equation form from parameter and force--motion consistency~\cite{wang2026gauge}.

\paragraph{From evaluation to diagnosis and learning.}
RigidBench intervenes on position representations~\cite{jain2026rigidbench}; WorldModelBench and PISA connect evaluation to reward-based post-training~\cite{li2025worldmodelbench,li2025pisa}. Our SI residuals and paired interventions distinguish coordinate, integration, and force-response errors within closed-chain motion.

\section{Measuring Mechanical Consistency}
\label{sec:audit}
\subsection{Physical coordinates and admissible acceleration}
We derive the audit for a planar chain with fixed ground, one driven crank, and independent coordinate $\theta=q_1$; Section~\ref{sec:spatial} extends it to space. For 4R chains, $q=(q_1,q_2,q_3)$ are absolute link angles, $(l_1,l_2,l_3)$ are link lengths, and $d$ is anchor spacing. With $e(\psi)=(\cos\psi,\sin\psi)\T$, closure is
\begin{equation}
\phi(q)=l_1e(q_1)+l_2e(q_2)-l_3e(q_3)-(d,0)\T=0.
\label{eq:closure}
\end{equation}
For a crank-slider, $q_3$ is the slider position $s$ and the last two terms become $-(s,h_s)\T$, where $h_s$ is guide offset. Crank-rocker and double-crank geometries share the 4R topology.

The physical audit uses hinges $B,C$ and the coupler midpoint $P$ as $x=(B,C,P)\in\R^6$. With fixed anchors $A,D$, the five point constraints are
\begin{equation}
\begin{aligned}
\Phi(x)=\bigl(&\|B-A\|-l_1,\ \|C-B\|-l_2,\\
&\|C-D\|-l_3,\ [P-(B+C)/2]\T\bigr)\T.
\end{aligned}
\label{eq:pointconstraints}
\end{equation}
Five constraints leave one degree of freedom in $\Phi(x)=0$. For sliders, guide height replaces the third constraint. Calibrated readouts give meter-valued $x$; rod directions give angles through $\operatorname{atan2}(r_z,r_x)$, and $C_x$ gives slider position. Raw closure is checked before conversion. Public geometry and camera calibration define the interface; independent mass, inertia, and friction define the evaluator.

Let $J=\partial\phi/\partial q$ and $\gamma=\partial q/\partial\theta$, with $J\gamma=0$, $\gamma_1=1$, and $\gamma'=\partial\gamma/\partial\theta$. Regular configurations obey
\begin{equation}
\dot q=\gamma\omega,\qquad
\ddot q=\gamma\alpha+\gamma'\omega^2,
\quad \omega=\dot\theta,\ \alpha=\ddot\theta.
\label{eq:admissible}
\end{equation}
The two terms change crank speed and follow path curvature, respectively. Force balance determines tangent acceleration $\alpha$.

\paragraph{Constraint preservation needs a correct initial level.}
KeyCLD's acceleration-level enforcement gives
\begin{equation}
D\Phi\,\ddot x+D^2\Phi[\dot x,\dot x]=0
\ \Longrightarrow\
\Phi(x(t))=\Phi(x_0)+tD\Phi(x_0)\dot x_0.
\label{eq:levelset}
\end{equation}
Velocity projection removes the second term. Acceleration-level enforcement then preserves the initial closure residual, including incorrectly decoded rod lengths, separating coordinate anchoring from subsequent dynamics.

\subsection{Independent rigid-body forces}
Reference rigid-body properties determine the independent mass matrix $M$, velocity bias $c$, and gravity $G$. Application-point Jacobians map external forces, while $Q_f$ combines viscous and smooth Coulomb friction in relative joint velocities (supplementary derivation).

With $e_1=(1,0,0)\T$ selecting the driven crank and $\lambda$ denoting constraint reactions, the full equation is
\begin{equation}
M\ddot q+c+G=e_1\tau_A+Q_{\rm ext}+Q_f+J\T\lambda.
\label{eq:dynamics}
\end{equation}
Classical reduced dynamics eliminates workless joint reactions by tangent projection. Because $J\gamma=0$, projection onto $\gamma$ exposes the predicted crank-torque mismatch:
\begin{equation}
r_\tau=\gamma\T\!\left[M\widehat{\ddot q}+c+G-Q_{\rm ext}-Q_f\right]-\tau_A.
\label{eq:torque}
\end{equation}
All terms use the predicted state; $\gamma_1=1$ fixes the crank-torque scale. The effective inertia $\mathcal I=\gamma\T M\gamma$ measures resistance to crank acceleration (supplementary derivation).

\subsection{Six complementary physical criteria}
The independent energy is $E=\tfrac12\dot q\T M\dot q+\sum_i m_i g e_z\T p_i$, where $p_i$ is body $i$'s center of mass and $e_z$ points upward. The generalized force $Q=e_1\tau_A+Q_{\rm ext}+Q_f$ collects motor, external, and friction forces; $Q\T\dot q$ is their total power. The work--energy residual compares energy change with supplied work:
\begin{equation}
r_E(t)=E(t)-E(t_0)-\int_{t_0}^t Q\T\dot q\,\dd t.
\label{eq:energy}
\end{equation}
For admissible motion, $r_E(t)=\int_{t_0}^t r_\tau(s)\omega(s)\,\dd s$. Torque tests local force balance; energy accumulates its work along the motion. At rest, power vanishes even with unbalanced forces.

The six criteria separate admissibility, mechanics, and tracking. Closure uses maximum absolute $\Phi(\hat x)$; tangent velocity and curvature-consistent acceleration use maximum-channel RMS of $\dot\Phi=D\Phi\hat v$ and $\ddot\Phi=D\Phi\hat a+D^2\Phi[\hat v,\hat v]$. Force and energy errors are RMS $r_\tau$ and maximum absolute prefix $r_E$. Finally, maximum-channel RMS of $\hat x-x^*$ tests whether the motion follows the reference trajectory, since an admissible path can still have the wrong phase. All use future samples, with $\hat v,\hat a$ supplied by the native ODE.

\begin{table}[t]
\caption{Engineering targets fixed by task and shared by its models. Torque/servo columns are planar; spatial effort and work scales follow Section~\ref{sec:spatial}.}
\label{tab:criteria}
\centering\small
\setlength{\tabcolsep}{3pt}
\begin{tabular}{lrrr}
\toprule
Criterion & Torque & Servo & Spatial\\
\midrule
Closure & $1$ mm & $1$ mm & $1$ mm\\
Velocity & $0.01$ m/s & $0.01$ m/s & $0.01$ m/s\\
Acceleration & $0.1$ m/s$^2$ & $0.1$ m/s$^2$ & $0.1$ m/s$^2$\\
Actuator RMS & $0.1$ N\,m & $0.25$ N\,m & $5\%$ rated\\
Energy defect & $0.02$ J & $0.05$ J & $1\%$ work\\
Point RMSE & $1$ cm & $1$ cm & $1$ cm\\
\bottomrule
\end{tabular}
\end{table}

\begin{figure*}[t]
\centering
\includegraphics[width=\textwidth]{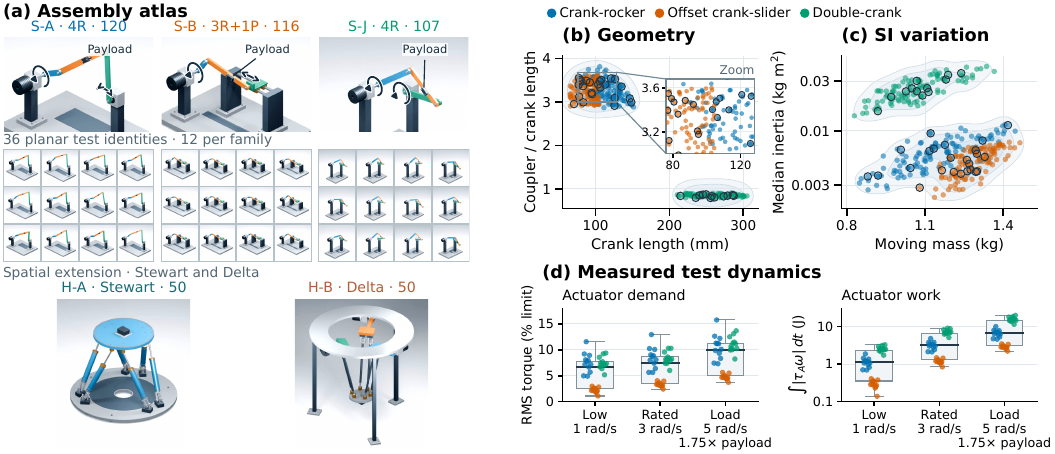}
\caption{Assembly corpus. (a) Three planar families and their 36 test identities, with Stewart/Delta examples. (b--d) Planar distributions: density contours cover 343 assemblies, test identities are ringed, and inertia is median over 41 poses. Torque and absolute work use 36 references per condition; boxes/whiskers show quartiles/ranges, and torque is normalized by actuator limits.}
\label{fig:mechanisms}
\end{figure*}

\paragraph{Continuous scoring at fixed engineering targets.}
The tolerances in Table~\ref{tab:criteria} express task requirements and are fixed before model evaluation. For case $i$ and criterion $k$, let $m_{ik}=1$ when its complete residual is measurable and $\rho_{ik}=r_{ik}/\tau_{ik}$ be its error-to-tolerance ratio. We use one continuous composite, with strict joint qualification as an auxiliary:
\begin{equation}
\begin{aligned}
S_{\rm comp}&=\frac{100}{6}\sum_{k=1}^6\left\langle\frac{m_{ik}}{1+\rho_{ik}^2}\right\rangle_{\rm bal},\\
S_{\land}&=100\left\langle\prod_{k=1}^6 m_{ik}\,\mathbf{1}[\rho_{ik}\le1]\right\rangle_{\rm bal}.
\end{aligned}
\label{eq:scores}
\end{equation}
Each criterion scores 100 at zero error, 50 at tolerance, and 20 at twice tolerance. Equal criterion weights allocate 50 points to kinematics, $33\frac{1}{3}$ to force/energy, and $16\frac{2}{3}$ to tracking. Component scores expose this allocation and its changes. Missing terms contribute zero in the fixed six-term denominator. We average trajectories within conditions, then conditions, families, and seeds equally. Coverage counts complete six-criterion measurements; $S_{\land}$ counts joint target attainment. Supplementary Table~\ref{app:thresholds} rescales all targets by $0.5/1/2$ on the saved residuals with measurement availability fixed.

\paragraph{Calibrating the measurement path.}
Reference and predicted RGB or features pass through the same frozen reader, reconstruction, and differentiator. Exact-state replay isolates numerical error: quintic splines pass all six criteria on 54/54 planar references. Correct-video replay adds localization error. Closure checks supplied lengths, while localization checks reference coordinates; both are measured at each resolution against fixed task targets.

\section{Experimental Design}
\label{sec:design}
The 30 Hz planar study separates resolution, anchoring, and integration (Section~\ref{sec:results}); 100 Hz torque excitation tests constitutive response (Section~\ref{sec:structure}); spatial controls compare predictive features and generated video (Section~\ref{sec:spatial}). Supplementary Sections~\ref{app:corpus-data}--\ref{app:geometry-readout} detail planar controls, Section~\ref{app:spatial} spatial protocols, and Sections~\ref{app:measurement}--\ref{app:resolution} measurement calibration and resolution.

\subsection{Shared assembly corpus}
MuJoCo~\cite{todorov2012mujoco} trajectories rendered in Blender span 343 crank-rockers, offset sliders, and double-cranks with varied geometry and rigid-body properties (\cref{fig:mechanisms}). We partition them into 271 training, 18 validation, 18 calibration, and 36 test identities, keeping each assembly's conditions and views together. Three conditions yield 108 test sequences per seed.

Both video models train at $64/256$ pixels from the same renders, with a $1.30$ m field of view and shared splits, motions, commands, and targets. Three observed frames start at $1$ s; training uses 50-frame windows and forecasting ends at $3.967$ s. Commands are $1$, $3$, or $5$ rad/s, with $1.75\times$ payload at the highest speed. The public controller maps predicted crank velocity to torque:
\begin{equation}
\tau_A=\operatorname{clip}[k_p(\omega_{\rm cmd}-\widehat\omega),-\tau_{\max},\tau_{\max}].
\label{eq:servo}
\end{equation}
Crank-rockers and sliders use $k_p=2$ N\,m\,s/rad and $\tau_{\max}=5\,\Nm$; double-cranks use $k_p=12$ N\,m\,s/rad and $\tau_{\max}=12\,\Nm$. The common $0.25\,\Nm$ criterion specifies absolute precision across these actuator scales. Forecasts use observed prefixes and public commands; future measured torque and reference states enter only the evaluator.

\subsection{Native objectives and diagnostic controls}
Models retain native architectures and objectives with shared-corpus interfaces. Readout interventions hold forecasts fixed; state controls supply exact initialization; paired training integrators share initialization, data order, objective, and budget.

KeyCLD combines image/heatmap reconstruction with four-step constrained prediction. CON jointly trains its CNN $\beta$-VAE and eight-coordinate oscillators over 50-frame windows~\cite{stolzle2024con}. Figure~\ref{fig:audit}b defines their respective torque/command interfaces.

\begin{figure*}[t]
\centering
\includegraphics[width=\textwidth]{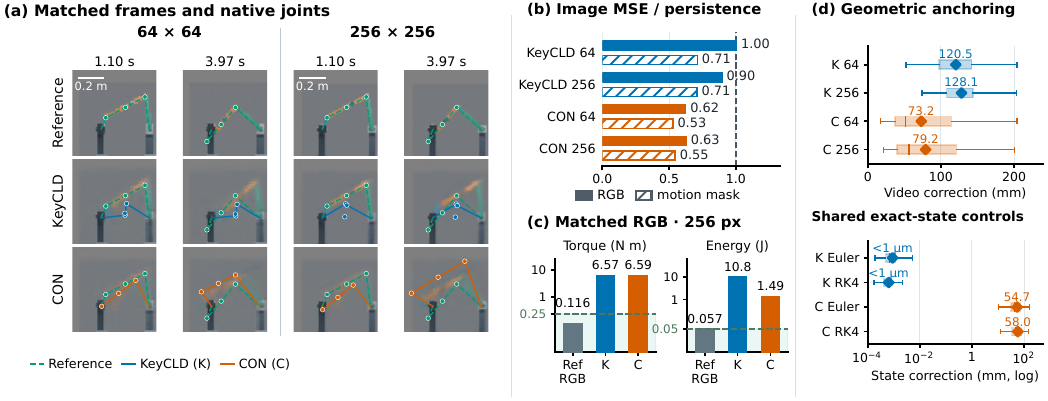}
\caption{Matched resolution and measurement controls. (a) The first test case, seed 42, at both resolutions; dashed/solid skeletons show reference/native joints. (b) Labeled mean MSE ratios to same-resolution persistence; solid/hatched bars show RGB/motion mask, and lower is better. (c) Correct and generated RGB through one frozen reader and phase differentiator; dashed lines mark SI targets. (d) Reconstruction costs: diamonds show balanced means, boxes quartiles, whiskers full ranges.}
\label{fig:shared-video}
\end{figure*}

\begin{table*}[t]
\caption{Planar audit: 36 assemblies $\times$ three conditions $\times$ three seeds per row. Finite / Phys. / Cov. count finite outputs / native SI coverage / reconstructed SI coverage. G/V/A are closure/velocity/acceleration pass rates; N/A marks unavailable mechanics. $S_{\rm comp}$ is mean$\pm$seed SD. Correction is coordinate RMS. Kinematics supply 50.00 of KeyCLD's 53.10 reconstructed points at 256 pixels. Point errors use maximum-channel RMSE on the native 240 Hz grid (Table~\ref{app:metric-paths}), with available-case means balanced by seed/family/condition. State controls are resolution-independent. Persistence RGB MSE is 4.94/5.04 at 64/256 pixels.}
\label{tab:corpus}
\label{tab:geometry-readout}
\centering\small
\setlength{\tabcolsep}{1.7pt}
\begin{tabular}{@{}lrrrrrrrrr@{\hspace{5pt}}rrrrr@{}}
\toprule
& \multicolumn{9}{c}{Native output: primary forecast assessment} & \multicolumn{5}{c}{Diagnostic geometric reconstruction}\\
\cmidrule(lr){2-10}\cmidrule(l){11-15}
Training & Finite & \shortstack{RGB MSE\\($10^{-4}$)} & Phys. & \shortstack{G / V / A\\(\%)} & \shortstack{Torque\\(\%)} & \shortstack{Energy\\(\%)} & \shortstack{Point\\(mm)} & \shortstack{Joint\\(\%)} & \shortstack{$S_{\rm comp}$\\(0--100)$\uparrow$} & Cov. & \shortstack{Correction\\(mm)} & \shortstack{Torque\\(N\,m)} & \shortstack{Energy\\(J)} & \shortstack{Point\\(mm)}\\
\midrule
\multicolumn{15}{l}{\textit{KeyCLD}}\\
Video 64 px & 311 & 4.96 & 0 & 0/22.2/65.1 & N/A & N/A & 229.2 & 0 & $15.73\pm5.00$ & 311 & 120.49 & 6.383 & 0.921 & 194.80 \\
Video 256 px & 324 & 4.52 & 0 & 0/53.7/75 & N/A & N/A & 233.9 & 0 & $22.30\pm12.08$ & 324 & 128.13 & 6.229 & 1.201 & 198.19 \\
State--Euler & 324 & -- & 324 & 100/100/100 & 8.33 & 0.00 & 139.1 & 0 & $52.65\pm0.02$ & 324 & $<0.001$ & 4.564 & 22.453 & 139.06 \\
State--RK4 & 324 & -- & 324 & 100/100/100 & 7.41 & 0.00 & 16.4 & 0 & $64.30\pm0.07$ & 324 & $<0.001$ & 0.715 & 2.925 & 16.37 \\
\midrule
\multicolumn{15}{l}{\textit{CON}}\\
Video 64 px & 324 & 3.08 & 0 & 0/0.9/33 & N/A & N/A & 123.5 & 0 & $7.23\pm3.23$ & 308 & 73.18 & 6.859 & 17.960 & 154.83 \\
Video 256 px & 324 & 3.17 & 0 & 0/0/63.6 & N/A & N/A & 134.6 & 0 & $10.96\pm2.50$ & 311 & 79.17 & 6.691 & 10.470 & 167.35 \\
State--Dopri5 & 324 & -- & 0 & 0/0/3.4 & N/A & N/A & 142.4 & 0 & $1.91\pm0.79$ & 323 & 67.68 & 6.549 & 23.282 & 132.28 \\
State--Euler & 324 & -- & 0 & 0/0/4.3 & N/A & N/A & 124.5 & 0 & $2.36\pm0.46$ & 317 & 54.73 & 6.028 & 11.847 & 133.06 \\
State--RK4 & 324 & -- & 0 & 0/0/3.4 & N/A & N/A & 137.4 & 0 & $1.97\pm0.84$ & 322 & 58.01 & 6.541 & 20.476 & 132.86 \\
\bottomrule
\end{tabular}
\end{table*}

At each resolution, KeyCLD completes 40 epochs and CON 50, giving 32,520 and 40,650 sequence exposures. Batch sizes of 1 and 16 yield 32,520 and 2,550 updates, respectively. The supplement details native image-layer scaling and pixel-loss normalization.

A validation-selected permutation maps KeyCLD keypoints to joints, while CON uses a training-label affine map. We freeze these readouts for evaluation and isolate dynamics with exact-state controls that supply true training states and prefix velocities.

Seeds 42--44 give twelve video fits and fifteen state fits at fixed final epochs. Confidence intervals resample seeds and assembly identities within families 2,000 times, retaining all conditions and paired arms. Native-derivative calibration passes 54/54 references and rejects static and time-scaled controls.

\begin{figure*}[t]
\centering
\includegraphics[width=\textwidth]{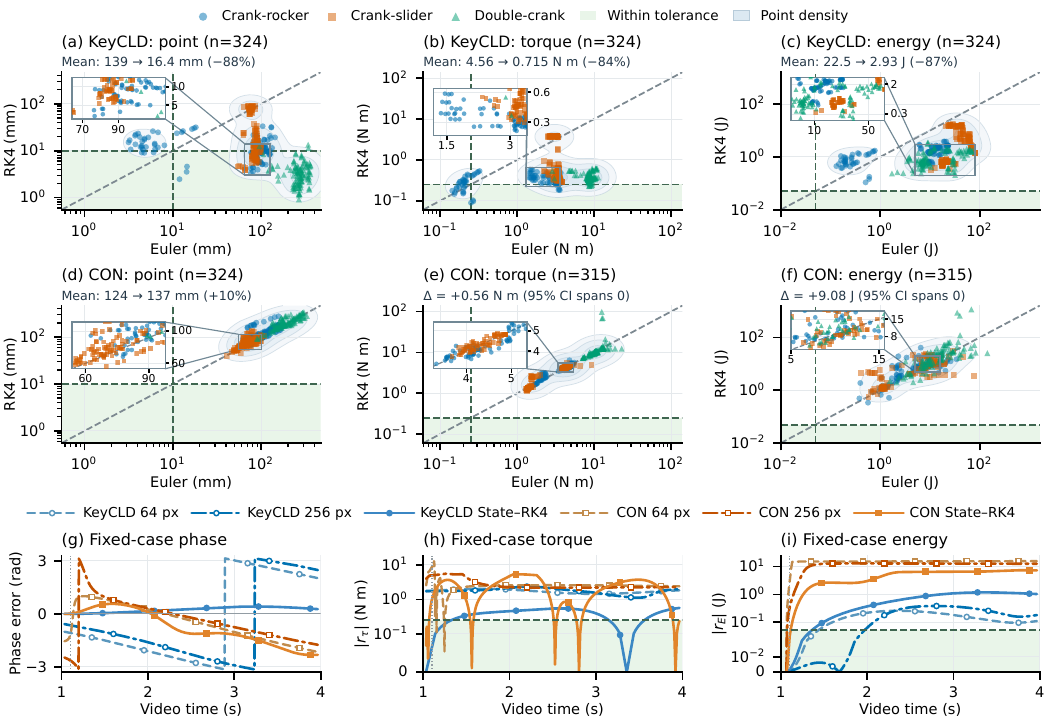}
\caption{Integration and resolution controls. (a--f) Resolution-independent Euler/RK4 pairs compare native point error and reconstructed force/energy; below-diagonal points favor RK4. (g--i) Six curves share one case, seed 42, time window, and geometric readout. Green bands mark tolerances; vertical dotted lines start forecasting.}
\label{fig:corpus-residuals}
\end{figure*}

\section{Shared Assembly Results}
\label{sec:results}
\subsection{Resolution intervention}
Retaining CON's native fully connected image layers expands its parameter count from 67.3M to 1.08B at $64\!\to\!256$ pixels, with 2,550 updates at either resolution. Its RGB gains over persistence are $37.6\%$ and $37.2\%$, with motion-region gains at both resolutions (Fig.~\ref{fig:shared-video}a,b). KeyCLD moves from near persistence to a $10.4\%$ mean RGB gain, and all 256-pixel forecasts are finite. These visual gains coexist with failed native geometry in all four settings (Table~\ref{tab:corpus}), separating visual prediction from physical coordinate anchoring.

Velocity and acceleration consistency drive continuous-score gains from 15.73 to 22.30 for KeyCLD and from 7.23 to 10.96 for CON, while point scores remain below 1/100 at 256 pixels. KeyCLD's seed SD reaches 12.08 (Table~\ref{tab:corpus}).

\paragraph{A common RGB measurement path.}
Frozen RGB readers reduce correct-video Euclidean error from 2.13 to 0.334 mm at $64\!\to\!256$ pixels, while 256-pixel forecasts give 195.0 mm for KeyCLD and 208.5 mm for CON. The shared projected-phase path gives point errors of 0.144, 183.0, and 181.9 mm for reference video, KeyCLD, and CON, respectively; reference $S_{\rm comp}$ is 90.00 with 66/108 joint passes (matched mechanics: Table~\ref{tab:paired-replay}; grids: Table~\ref{app:metric-paths}). The augmented 64-pixel reader's 2.73 mm mean error under synthetic blur and reduced contrast (Table~\ref{app:reader-quality}) also lies well below the measured forecast drift.

\subsection{What changes when the state is known?}
Exact-state KeyCLD passes all three kinematic criteria on 324/324 forecasts under both Euler and RK4 training. RK4 reduces point error from 139.1 to 16.4 mm, torque error from 4.564 to 0.715 N\,m, and energy error from 22.453 to 2.925 J. Tracking (+7.81 points) and torque (+3.82) drive an 11.65-point $S_{\rm comp}$ gain (paired 95\% CI $[10.87,12.48]$; Table~\ref{app:state-components}). Binary torque qualification is 8.33\% for Euler and 7.41\% for RK4, with zero energy passes for either. The score advantage persists at half and twice the targets (Table~\ref{app:thresholds}).

CON's 13,360-parameter state model evolves six Cartesian coordinates padded to eight oscillator coordinates. Its forecasts pass geometry on 0/324 cases with exact-state supervision; RK4 increases point error by $12.907$ mm (95\% CI $[6.734,23.497]$). Integration changes trajectory accuracy in opposite directions in these two state pipelines.

\subsection{Force and energy after a common geometric readout}
Predicted crank direction and derivatives define a causal reconstruction using public geometry. Video corrections span 73.18--128.13 mm across models and resolutions; exact-state KeyCLD needs less than a micrometer. Reconstruction raises 256-pixel $S_{\rm comp}$ to 53.10 for KeyCLD and 48.26 for CON. These reconstructed video trajectories retain torque and energy residuals at least 24 times their targets, exposing force and work imbalances within kinematically valid motion (Table~\ref{tab:geometry-readout}).

\section{Structural and Intervention Analysis}
\label{sec:structure}
\begin{figure*}[t]
\centering
\includegraphics[width=\textwidth]{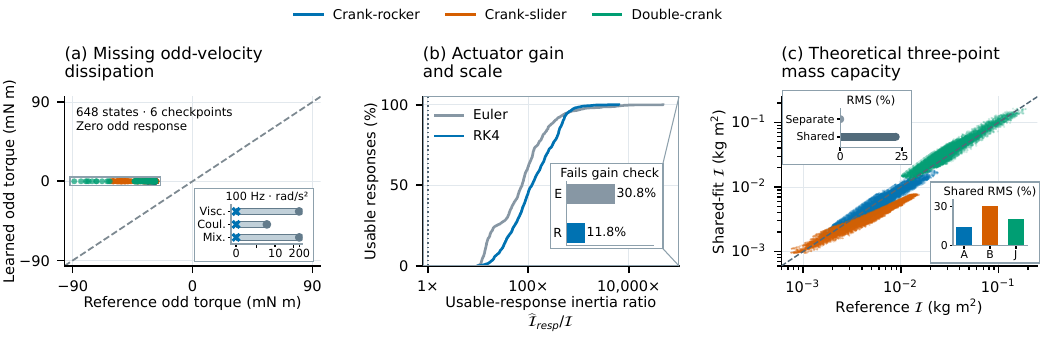}
\caption{Structural evidence. (a) Six checkpoints give zero odd torque on 648 states. The 100 Hz inset compares RMS odd acceleration (reference bars/dots, model crosses; symlog scale). (b) Each integrator pools 1,944 queries; the inset shows failed gain checks. (c) Mass fits use 44,280 poses, with separate/shared and family errors in the insets.}
\label{fig:constitutive}
\end{figure*}

KeyCLD's torque interface lets us test whether improved rollouts also yield calibrated force response. Velocity reversal, torque perturbation, and mass-capacity fits distinguish dissipation, actuator scale, and cross-assembly parameter sharing (Fig.~\ref{fig:constitutive}).

\subsection{Torque excitation: dissipation and calibration}
The torque-excitation protocol uses one training seed and direct $[-2,2]\,\Nm$ inputs on a frictionless four-bar at 100 Hz. Eight friction, load, scale, and topology conditions provide 96 calibration and 160 transfer trajectories. With known geometry and calibrated torque, exact-state regression has rank 12/12 and recovers masses to $1.33\times10^{-6}$ maximum relative error across these conditions (Table~\ref{app:identification}).

At fixed pose and torque, KeyCLD's force $f(x,u)=-\nabla V(x)+B(x)u$ is velocity-independent and its constraint-curvature term is quadratic in velocity. Its acceleration is therefore unchanged by reversal. The half-difference isolates friction's odd response, cancelling gravity and other velocity-even terms:
\begin{equation}
\frac{\alpha(\theta,\omega,\tau)-\alpha(\theta,-\omega,\tau)}{2}
=-\frac{\mathcal F(\theta,\omega)}{\mathcal I(\theta)}.
\label{eq:odd}
\end{equation}
All 292 reversal pairs per friction condition have zero learned odd acceleration. The nonzero viscous, Coulomb, and mixed-friction references in Fig.~\ref{fig:constitutive}(a) expose a representational restriction, consistent with the missing dissipation identified by KeyCLD~\cite{daems2024keycld}.

Rescaling $(M_x,V,B)\mapsto(kM_x,kV,kB)$, $k>0$, preserves acceleration, so motion fitting alone cannot identify absolute SI parameter scales. We test acceleration under a calibrated torque perturbation at fixed pose and zero speed:
\begin{equation}
k_\tau=\frac{\hat\alpha(+\delta\tau)-\hat\alpha(-\delta\tau)}{2\delta\tau},
\qquad \widehat{\mathcal I}_{\rm resp}=1/k_\tau.
\label{eq:gain}
\end{equation}
The physical gain $k_\tau=1/\mathcal I$ tests the combined SI calibration of inertia and actuation.

\begin{figure*}[t]
\centering
\includegraphics[width=\textwidth]{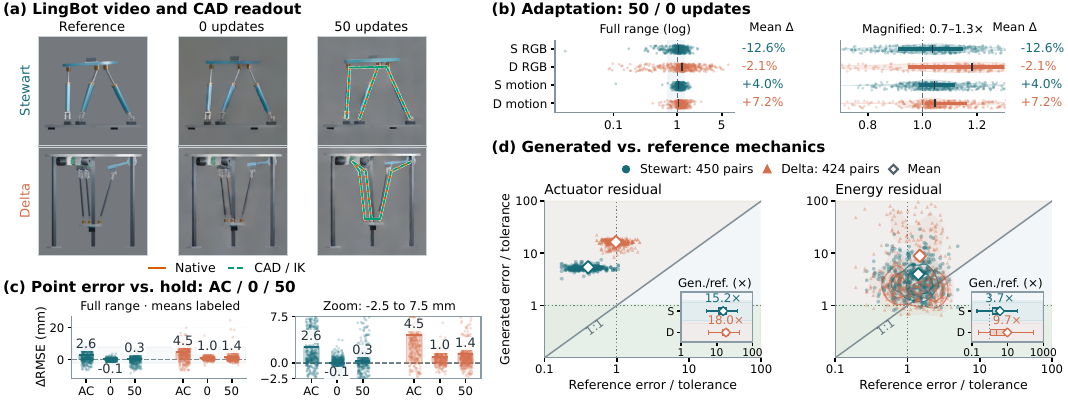}
\caption{Spatial diagnostics. (a) LingBot video and CAD readouts. (b) Paired 50/0-update MSE ratios. (c) Finite point errors and mean bars (450 registered per setting/family). (d) Mechanics on 450 Stewart and 424 Delta pairs. Warm/green regions mark errors above reference/within tolerance; contours show density, diamonds means, and insets paired-ratio quartiles, ranges, and means.}
\label{fig:spatial}
\end{figure*}

\subsection{Shared assemblies: response and representation}
Each of the six frozen 30 Hz KeyCLD state checkpoints is queried at the same 648 held-out states. Holding torque fixed removes servo feedback that can resemble damping; the learned odd response remains zero. With $\delta\tau=0.25\,\Nm$ and three pooled seeds, failures of the finite, $k_\tau>10^{-8}$ gain check fall from $30.8\%$ (Euler) to $11.8\%$ (RK4). Usable responses still have median inertia ratios of $74.9\times$ and $117.3\times$ the reference. The fixed-state probe thus reveals incorrect response scaling despite RK4's lower rollout residuals.

Three-point reduced inertia is
\begin{equation}
\mathcal I_{\rm pts}(\theta)=\sum_{i\in\{B,C,P\}}m_i
\left\|\frac{\partial p_i}{\partial\theta}\right\|^2.
\label{eq:capacity}
\end{equation}
Positive-mass fits across 360 assemblies reach numerical precision separately by condition; one shared mass vector incurs $22.55\%$ relative RMS error (Fig.~\ref{fig:constitutive}c). Cross-mechanism and load variation requires conditioned masses.

\section{Predictive and Generative Spatial Dynamics}
\label{sec:spatial}
On Stewart and Delta, both models forecast 117 future frames at 30 Hz on the same 50+50 test assemblies, three conditions, and seeds 42--44. V-JEPA trains its 24-layer AC predictor on frozen 384-pixel encoder features for 3,000 exposures per family and seed; its internal state head supplies autoregressive feedback. LingBot generates $256\times256$ RGB at 0 and 50 full-parameter updates (400 exposures per seed), giving 1,800 videos. Fixed measurement heads decode unconstrained (native) coordinates, with matched reference replay establishing each pathway's measurement baseline.

For either pathway, a tangent basis $N$ represents the mechanism's permitted motions. With actuator map $B$, the reaction-free residual becomes
\begin{equation}
r_u=(N\T B)^{-1}N\T[M\hat a+c+G-Q_{\rm ext}-Q_f]-u_A.
\label{eq:spatial-effort}
\end{equation}
Here $JN=0$ eliminates internal reactions, and nonsingular $N\T B$ converts the imbalance to actuator units. Controllers are reevaluated at predicted states. Effort targets are 7.5 N for Stewart and 0.75 N\,m for Delta; the energy target is $0.01u_{\max}\sum_j\operatorname{range}(q_{{\rm cmd},j})$, spanning 0.106--0.504 J over these trajectories.

Reference replay establishes mean closure errors of 2.41 mm for Stewart and 3.12 mm for Delta against the 1 mm target; native AC readouts have zero mechanical coverage. With the same frozen calibrated head and CAD reconstruction, forecast effort residuals are 13.5 times reference residuals on Stewart and 6.0 times on Delta (Table~\ref{tab:paired-replay}). Decoding saved Stewart forecasts with the calibrated measurement head in place of the internal feedback head raises CAD $S_{\rm comp}$ from 43.51 to 55.11 (Table~\ref{app:ac-readout-table}), isolating the contribution of coordinate recovery at fixed latent trajectories.

\begingroup
\setlength{\intextsep}{5pt}
\begin{table}[H]
\caption{Spatial audit, 50 assemblies $\times$ three conditions $\times$ three seeds per row. Step counts updates. Scores are mean$\pm$seed SD; Cov. counts complete CAD measurements. RGB/motion MSE units: $10^{-4}/10^{-2}$.}
\label{tab:spatial}
\centering\footnotesize
\setlength{\tabcolsep}{2.0pt}
\begin{tabular}{@{}lrrrrrr@{}}
\toprule
& & \multicolumn{3}{c}{Native output} & \multicolumn{2}{c}{CAD intervention}\\
\cmidrule(lr){3-5}\cmidrule(l){6-7}
Family & Step & RGB & Motion & $S_{\rm comp}$ & $S_{\rm comp}$ & Cov.\\
\midrule
Stewart & 0 & 41.67 & 2.59 & $5.67\!\pm\!0.05$ & $54.60\!\pm\!0.19$ & 450/450 \\
Stewart & 50 & 36.42 & 2.69 & $6.01\!\pm\!0.15$ & $54.31\!\pm\!0.51$ & 450/450 \\
Delta & 0 & 29.43 & 1.58 & $2.80\!\pm\!0.08$ & $43.37\!\pm\!1.17$ & 359/450 \\
Delta & 50 & 28.83 & 1.70 & $3.39\!\pm\!0.02$ & $51.92\!\pm\!1.17$ & 424/450 \\
\midrule
AC Stewart & 386 & -- & -- & $6.20\!\pm\!0.69$ & $55.11\!\pm\!1.71$ & 450/450 \\
AC Delta & 400 & -- & -- & $10.40\!\pm\!5.34$ & $50.05\!\pm\!2.71$ & 432/450 \\
\bottomrule
\end{tabular}
\end{table}
\endgroup

LingBot adaptation lowers whole-image MSE by 12.6\% on Stewart and 2.1\% on Delta, while motion-region MSE rises by 4.0\% and 7.2\%, respectively (Fig.~\ref{fig:spatial}b). Thus the global visual gain accompanies reduced foreground accuracy. Both checkpoints exceed image-persistence error and have zero native mechanics coverage.

At 50 updates, CAD/IK reconstruction gives complete measurements on 450/450 Stewart and 424/450 Delta forecasts. Corrections average 2.11 and 9.06 mm, with point errors of 29.92 and 23.24 mm, respectively. Delta's score rises from 43.37 to 51.92 through coverage and kinematic gains, while Stewart stays near 54 and joint qualification remains zero.

Matched reference replay (Table~\ref{tab:paired-replay}) shows larger generated effort on all 450 Stewart and 424 Delta pairs, with larger energy defects on 390 and 330 pairs, respectively. The held CAD reset yields zero energy defect but 12.13 N\,m torque error on six Delta clips (Table~\ref{app:delta-components}): at zero velocity, unbalanced effort contributes zero power.

\begin{table}[t]
\caption{Matched reference replay and prediction. Each pair uses identical cases, a frozen reader, and the same reconstruction and derivatives. Planar rows use 256-pixel RGB with projected phase; spatial rows use CAD. Effort is N for Stewart and N\,m otherwise; energy is J. LingBot uses 50 updates. Reference energy passes on Stewart and Delta are 150/150 each for exact and 30 Hz sampled states, versus 48/150 and 63/150 through RGB (Table~\ref{app:spatial-controls}).}
\label{tab:paired-replay}
\centering\footnotesize
\setlength{\tabcolsep}{2.5pt}
\begin{tabular}{@{}lrrrrr@{}}
\toprule
& & \multicolumn{2}{c}{Effort RMS} & \multicolumn{2}{c}{Energy defect}\\
\cmidrule(lr){3-4}\cmidrule(l){5-6}
Path / prediction & $n$ & Ref. & Pred. & Ref. & Pred.\\
\midrule
Planar / KeyCLD & 324 & 0.116 & 6.565 & 0.057 & 10.835\\
Planar / CON & 324 & 0.116 & 6.585 & 0.057 & 1.485\\
\midrule
AC / Stewart & 450 & 3.108 & 41.967 & 0.627 & 1.504\\
AC / Delta & 432 & 2.083 & 12.425 & 0.985 & 2.462\\
\midrule
LingBot / Stewart & 450 & 3.009 & 40.513 & 0.605 & 1.640\\
LingBot / Delta & 424 & 0.733 & 12.319 & 0.203 & 1.206\\
\bottomrule
\end{tabular}
\end{table}

\section{Implications and Conclusion}
\label{sec:discussion}
Reference-calibrated SI measurements and controlled interventions distinguish coordinate, numerical, and constitutive errors. Visual gains, geometric validity, and energy balance resolve different aspects of prediction; paired force probes expose response errors within improved rollouts. The resulting design requirements are coordinate anchors, velocity-dependent dissipation, calibrated SI actuator response, and parameters conditioned on geometry and load.
\par
\label{maintextend}
\clearfullpage
{\small
\bibliographystyle{ieeenat_fullname}
\bibliography{references}
}
\clearpage
\appendix
\setcounter{figure}{0}
\setcounter{table}{0}
\setcounter{equation}{0}
\renewcommand{\thefigure}{S\arabic{figure}}
\renewcommand{\thetable}{S\arabic{table}}
\renewcommand{\theequation}{S\arabic{equation}}
\renewcommand{\theHfigure}{supp.\arabic{figure}}
\renewcommand{\theHtable}{supp.\arabic{table}}
\renewcommand{\theHequation}{supp.\arabic{equation}}
\maketitlesupplementary
\pdfbookmark[0]{Supplementary Material}{supplement}
\label{supplementstart}

The supplement follows the main paper's progression from visual coordinates to admissible motion, actuator effort, and work--energy balance. Physical derivations define the measured quantities, reference replay calibrates their recovery, and controlled interventions distinguish the sources of error. The torque-excitation study isolates KeyCLD's force response; the shared planar corpus compares KeyCLD and CON; and the spatial study extends the same criteria to V-JEPA 2.1 and LingBot-VA on Stewart and Delta mechanisms. Section~\ref{app:measurement} details the observation and differentiation controls.

\section{Geometry and parameter specification}
\label{app:mechanism}
The planar mechanism lies in the world $XZ$ plane, with revolute axes along $-Y$. Ground anchors are $A=(0,0.32)$ m and $D=(0.40,0.32)$ m. Absolute link angles are measured in the motion plane. The network interface uses physical points $B,C$ and $P=(B+C)/2$. The five supplied constraints are the three distances $\lVert B-A\rVert=l_1$, $\lVert C-B\rVert=l_2$, $\lVert C-D\rVert=l_3$, together with the two midpoint coordinates. For the slider, $C_z=0.32$ m replaces the rocker distance. These constraints leave one independent degree of freedom in the six point coordinates.

Each moving link is a uniform capsule plate with two through bores. Its pivot spacing is $L$, width $2r=0.028$ m, thickness $h=0.012$ m, bore radius $r_b=0.005$ m, and density $\rho=2700$ kg/m$^3$. Its mass is
\begin{equation}
m=\rho h\left(2rL+\pi r^2-2\pi r_b^2\right).
\end{equation}
Analytic area moments give the inertia about the center of mass; the fixed ground and visual fasteners do not add unrecorded moving mass. The three moving masses are $0.123725$, $0.341453$, and $0.268877$ kg. The source payload is a $0.052\times0.012\times0.024$ m cuboid of mass $0.08$ kg, attached $0.24$ m along the coupler. Its local in-plane normal offset is zero. The rotor armature is $0.0002$ kg\,m$^2$, gravity is $9.81$ m/s$^2$, and a constant world force $(0,-0.5)$ N acts $0.30$ m along the coupler.

The known camera is 10 m from the plane, centered at $(0.20,0.37)$ m in $XZ$, with a field of view determined by a $0.43$ m half extent. The KeyCLD pixel-center convention gives the physical map
\begin{equation}
x_{\rm world}=0.43\frac{63}{64}x_{\rm normalized}
+(0.20,0.37).
\end{equation}
One fixed calibration serves all conditions, with small offsets in the visible thickness direction retained in rendering.

\begin{table}[t]
\caption{Changes from the nominal source condition. Unlisted physical parameters retain their source values.}
\label{app:conditions}
\centering
\small
\begin{tabular}{lp{.66\columnwidth}}
\toprule
Condition & Change\\
\midrule
Viscous & $b=(0.10,0.03,0.02,0.04)$ N\,m\,s/rad\\
Coulomb & $f_c=(0.008,0.005,0.005,0.007)$ N\,m\\
Mixed & Both preceding friction terms\\
Payload & Payload mass $0.25$ kg; cuboid size fixed\\
Density & Bar density $4050$ kg/m$^3$; payload fixed\\
Size & Link and ground lengths $\times0.8$; along-coupler force/payload offsets $\times0.8$; width and thickness fixed\\
Slider & Rocker replaced by a $0.055\times0.024\times0.044$ m slider of mass $0.268877$ kg with an ideal horizontal prismatic joint\\
\bottomrule
\end{tabular}
\end{table}

For the four-bar, actual relative joint velocities are $R\dot q$, where
\begin{equation}
R=\begin{bmatrix}
1&0&0\\-1&1&0\\0&-1&1\\0&0&1
\end{bmatrix}.
\end{equation}
With elementwise multiplication, signed friction is
\begin{equation}
Q_f=-R\T\left[b\odot R\dot q+
f_c\odot\tanh(R\dot q/\epsilon)\right],
\end{equation}
where $\epsilon=0.05$ rad/s, $b$ contains joint viscous coefficients, and $f_c$ contains smooth Coulomb torque magnitudes in SI units. This law models sliding friction; exact stiction and prismatic friction are outside the current model. The slider condition is frictionless.

\section{Independent dynamics and energy}
The crank-normalized tangent converts the constrained equations into one scalar balance:
\begin{equation}
\mathcal I\alpha+\mathcal H\omega^2+\mathcal G+\mathcal F=\tau_A+\mathcal Q_{\rm ext},
\label{eq:reduced}
\end{equation}
where $\mathcal I=\gamma\T M\gamma$, $\mathcal H=\gamma\T(M\gamma'+c(q,\gamma))$, $\mathcal G=\gamma\T G$, $\mathcal F=-\gamma\T Q_f$, and $\mathcal Q_{\rm ext}=\gamma\T Q_{\rm ext}$. For admissible motion, the local torque residual accumulates into the work--energy defect:
\begin{equation}
r_E(t)=\int_{t_0}^t r_\tau(s)\omega(s)\,\dd s.
\label{eq:residualpower}
\end{equation}
\label{app:equations}
For body $i$, let $p_i(q)$ be its center of mass, $J_i=\partial p_i/\partial q$, and $h_i=\dot J_i\dot q$. Its planar angular velocity is $s_i\T\dot q$, with constant selector $s_i$; a translating slider has $s_i=0$. From prescribed mass $m_i$, out-of-plane inertia $I_i$, rotor inertia $I_r$, upward unit vector $e_z$, and gravitational magnitude $g$, we construct
\begin{align}
M&=\sum_i\left(m_iJ_i\T J_i+I_i s_i s_i\T\right)+I_r e_1e_1\T,\\
c&=\sum_i m_iJ_i\T h_i,\qquad
G=\sum_i m_i gJ_i\T e_z.
\label{eq:bodydynamics}
\end{align}
Here $e_1=(1,0,0)\T$. The point-force Jacobian gives $Q_{\rm ext}=J_F\T F$. Relative joint velocities are $w=R\dot q$: for 4R joints, $w=(\dot q_1,\dot q_2-\dot q_1,\dot q_3-\dot q_2,\dot q_3)\T$. In a slider chain the last two entries are $-\dot q_2$ and $\dot s$. Viscous and smooth Coulomb forces are
\begin{equation}
Q_f=-R\T\!\left[b\odot w+f_c\odot\tanh(w/\epsilon)\right],
\label{eq:friction}
\end{equation}
where $b,f_c,\epsilon$ are channel-wise damping coefficients, Coulomb magnitudes, and smoothing velocities. Revolute and prismatic entries retain their respective SI units.

The independent tangent is obtained from
\begin{equation}
\gamma_1=1,\qquad
J_{:,2:3}\gamma_{2:3}=-J_{:,1}.
\end{equation}
Differentiating $J\gamma=0$ gives $\gamma'$ and hence
\begin{equation}
\dot q=\gamma\omega,\qquad
\ddot q=\gamma\alpha+\gamma'\omega^2.
\end{equation}
Projection of the full force equation then gives the reduced scalar acceleration used by the reference solver. The current configurations remain in an invertible crank chart. A multi-degree-of-freedom extension needs a registered physical chart and channel units, especially near singularities.

The score evaluator decodes raw point derivatives analytically. For a planar rod vector $r$, with point-derived derivatives $\dot r,\ddot r$,
\begin{align}
\dot\theta&=\frac{r_x\dot r_z-r_z\dot r_x}{\|r\|^2},\\
\ddot\theta&=\frac{r_x\ddot r_z-r_z\ddot r_x}{\|r\|^2}
-2\frac{(r\T\dot r)(r_x\dot r_z-r_z\dot r_x)}{\|r\|^4}.
\end{align}
Native scoring uses unprojected coordinates, with raw link-length and midpoint errors measured before this conversion. Section~\ref{app:geometry-readout} defines the separate geometric-readout intervention.

The energy computation uses physical kinetic energy and gravitational potential. Constant crank torque performs work $\tau_A[\theta(t)-\theta(t_0)]$, with the crank angle unwrapped on the dense grid. A constant world force performs work $F\T[p_F(t)-p_F(t_0)]$. Signed friction power $Q_f\T\dot q$ is integrated at 1600 Hz. Fixed ideal constraints perform no work. These identities avoid differentiating video positions twice or approximating the motor work with coarse quadrature.

\section{Training and diagnostic controls}
KeyCLD combines visual reconstruction, short-rollout prediction, and heatmap regularization:
\begin{equation}
\begin{aligned}
\mathcal L_{\rm video}&=\mathcal L_{\rm RGB}+\mathcal L_{\rm dyn}+\mathcal L_{\rm heat},\\
\mathcal L_{\rm dyn}&=\sum_t\sum_{k=0}^{4}\|\hat x_{t+k\mid t}-x_{t+k}\|^2.
\end{aligned}
\label{eq:learning}
\end{equation}
\label{app:training}
The full video adaptation follows the released KeyCLD modules and loss construction~\cite{daems2024keycld}. The pinned public revision is
\texttt{426c99a53d597e81044e4ce77214cd3aa6b92b64}.
There are 596,508 parameters in the adapted model. The encoder and renderer use base channel width 32, expanding to 64 and 128 features. The potential and input-map MLPs have two hidden layers of width 32 with CELU activation. The mass is represented by three squared point-mass parameters.

The three losses are image reconstruction, heatmap consistency, and four-step latent dynamics prediction. Dynamics and heatmap weights are one. Adam uses the released parameter groups, with visual learning rate $3\times10^{-4}$, dynamics learning rate $6\times10^{-5}$, and elementwise gradient clipping at 5. The returned optimizer state is carried forward each update, repairing its omission in the pinned update path. The mechanism geometry, three-point representation, recorded seed, and test sets specify the closed-chain adaptation of the released procedure.

The state-supervised controls replace image-derived point positions and finite-difference velocities in the dynamics loss with their true values. They use the same initial dynamics parameters, 450 training trajectories, 50 validation trajectories, batch size one, 40 epochs, and 18,000 updates. The visual networks are inactive and remain at initialization. The sole paired difference is Euler training integration versus four RK4 substeps per 0.01 s. Both final checkpoints use the same adaptive solver at test time.

\begin{table}[t]
\caption{State-control force diagnostic on 50 original test trajectories. This uses a \emph{unit-norm} null-space tangent $n$, alongside the crank-normalized $\gamma$ used for actuator effort.}
\centering
\small
\begin{tabular}{lrr}
\toprule
Control & Rod-length RMSE ($\mu$m) & Tangent force RMS\\
\midrule
\statee & 0.121 & 0.977704 N\,m\\
\stater & 0.449 & 0.459588 N\,m\\
\bottomrule
\end{tabular}
\end{table}

On the torque-driven held-out set, changing inference integration to RK4 with 16 substeps leaves VPT unchanged. The paired difference in future image MSE is $-1.88\times10^{-7}$, with 95\% bootstrap interval $[-5.61\times10^{-7},4.06\times10^{-9}]$. These checks isolate inference integration at fixed learned dynamics.

Confidence intervals for the torque-excitation diagnostic comparison use 2,000 paired trajectory-bootstrap resamples of the existing 50 cases. They condition on one training seed. The state-RK4 minus state-Euler zero-velocity acceleration error is $-0.70067$, with interval $[-0.73449,-0.66339]$. The corresponding unit-tangent force difference is $-0.51812$ N\,m, with interval $[-0.63616,-0.40859]$. Unit-tangent force and crank-equivalent torque use their respective normalizations.

\subsection{Torque-excitation visual and state diagnostics}
\begin{figure*}[t]
\centering
\includegraphics[width=\textwidth]{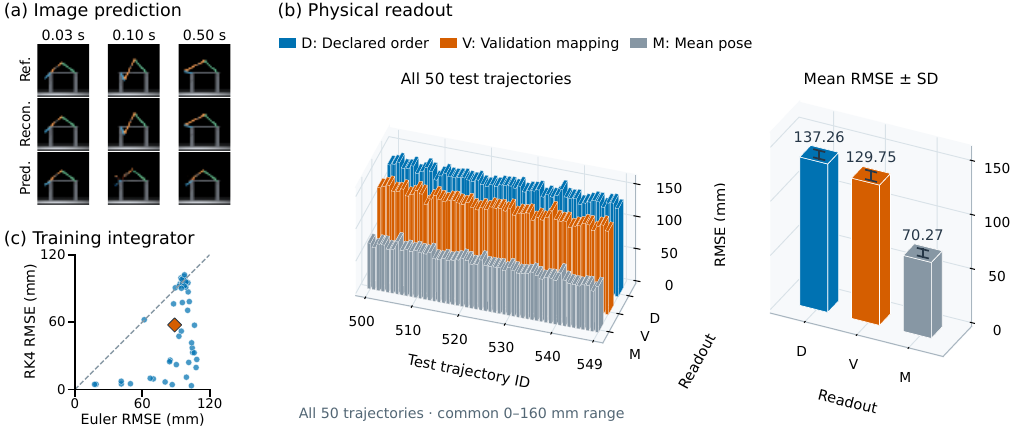}
\caption{Visual and integration diagnostics. (a) Reference, framewise reconstruction, and forecast at three fixed times. Reconstruction uses each true image; forecasts use three initial frames. (b) Fifty trajectories $\times$ three readouts; adjacent bars show mean coordinate RMSE $\pm$ SD (mm). (c) Paired state-supervised Euler/RK4 errors; diamond: mean.}
\label{fig:diagnostics}
\end{figure*}

On 50 torque-driven KeyCLD tests, framewise reconstruction and prefix forecasting give MSE $1.18\times10^{-4}$ and $8.94\times10^{-3}$ (\cref{fig:diagnostics}a). The gap remains over $70\times$ on motion pixels, which occupy $17.06\%$ of each image yet contain $95.74\%$ of forecast squared error, averaged per trajectory.

Video VPT is $7.82\pm4.86$, versus persistence's $7.98\pm4.17$: paired difference $-0.16$, 95\% interval $[-0.42,0.18]$. Resetting 44 local windows per trajectory to observed encodings gives four-step Euler latent RMSE $6.774$ mm versus hold's $6.657$ mm. The $+0.117$ mm difference, interval $[0.086,0.153]$, exposes a local prediction deficit.

A validation-selected keypoint permutation reduces coordinate RMSE from $137.26$ to $129.75$ mm ($5.47\%$), still above the mean-pose control's $70.27$ mm (\cref{fig:diagnostics}b). The small gain from relabeling indicates a coordinate-anchoring problem in the native objective. Encoded constraint RMSE is $158.53$ mm, yet its level drifts by only $0.039\,\mu$m: the dynamics preserve the miscalibrated initial geometry, as derived in the main text.

\subsection{Visual-region and reset-window diagnostics}
To separate moving mechanisms from static background, the evaluation mask marks pixels whose mean RGB temporal variance over all 50 reference images exceeds $10^{-4}$. With mask fraction $p_i$, the fraction of forecast squared error inside it is $s_i=p_i\,\mathrm{MSE}_{i,\mathrm{motion}}/\mathrm{MSE}_{i,\mathrm{full}}$. Computing these quantities per trajectory and then averaging the 50 tests gives $\langle p_i\rangle=0.17055$ and $\langle s_i\rangle=0.95738$. RGB errors use future frames 3--49 (zero-based), while the reconstruction/forecast comparison holds the visual encoder and renderer fixed.

\begin{table}[t]
\caption{Visual controls on the 50 torque-driven tests. MSE averages normalized RGB squared error. Motion MSE uses the temporal-variance mask. Framewise reconstruction receives each true image; the other controls use the observed prefix or a training-set statistic. VPT retains the author's index convention.}
\label{app:visual-controls}
\centering\small
\begin{tabular}{lrrr}
\toprule
Control & RGB MSE & Motion MSE & VPT\\
\midrule
Framewise recon. & 0.000118 & 0.000696 & 49.00 \\
Video forecast & 0.00894 & 0.0506 & 7.82 \\
Last-frame persistence & 0.00898 & 0.0533 & 7.98 \\
Training-mean image & 0.00483 & 0.0283 & 8.10 \\
\bottomrule
\end{tabular}
\end{table}

Each local window resets position to the observed encoding and velocity to its central difference; 44 windows per trajectory retain the same action and frozen checkpoint. Four-step coordinate RMSE averages $6.774$ mm for Euler, $6.796$ mm for RK4, and $6.657$ mm for latent persistence. The targets are image encodings in calibrated coordinate units. Post-hoc paired trajectory-bootstrap intervals use 2,000 draws with seed 20261013. Euler-minus-persistence is $0.117$ mm, with 95\% interval $[0.086,0.153]$; RK4-minus-persistence is $0.139$ mm, with interval $[0.104,0.176]$. Resampling trajectories retains dependence among their local windows.

The released RGB loss decodes observed keypoints, while its dynamics term supervises future encoded positions. Thus image reconstruction trains the representation locally, but neither future rendered images nor absolute closure residuals anchor its long-horizon coordinates. The numerical controls locate errors along this training pathway. On the assembly corpus, the $159.09$--$161.99$ mm point RMSEs describe validation-set permutation selection; held-out test coverage assesses the resulting frozen readout.

\section{Data stages and numerical calibration}
\label{app:calibration}
\begin{table}[t]
\caption{Disjoint data roles used in the paper. Transfer probes and final scoring use different test seeds.}
\centering
\small
\begin{tabular}{lrrl}
\toprule
Role & Conditions & Clips & Seeds\\
\midrule
Source training & 1 & 450 & 0--449\\
Source validation & 1 & 50 & 450--499\\
Original held-out test & 1 & 50 & 500--549\\
Extended transfer/probes & 8 & 160 & 1000--1019\\
Torque calibration & 8 & 96 & 2000--2011\\
Torque test & 8 & 160 & 3000--3019\\
\bottomrule
\end{tabular}
\end{table}

For the torque-excitation audit, a physical trajectory supplies initial images, action metadata, and a separately stored dense reference. The video model receives the three images at 0.01, 0.02, and 0.03 s. Its original central-difference initialization is centered at 0.02 s. The state component receives the true state and velocity at the same time. Each frozen model is integrated to 0.50 s and evaluated from 0.04 s onward, yielding 737 scored samples at 1600 Hz.

With ideal DOP853 references and endpoint work computation, all 96 positive controls pass. Maximum geometric error is $3.33\times10^{-16}$ m, maximum trajectory-level torque RMS is $3.45\times10^{-14}$ N\,m, and maximum energy residual is $3.49\times10^{-5}$ J. MuJoCo provides an independent mass/bias implementation check and the renderer. These controls calibrate the physical-coordinate interface independently of visual localization.

A separate diagnostic of the common finite-difference route on a sampled nominal trajectory gives torque RMS $2.439$ N\,m with a seven-point cubic differentiator at 100 Hz, even for the correct sampled state. It motivates separating that route from the native-ODE scoring track. The latter includes five-point dense differentiation checks against native derivatives and an 800 versus 1600 Hz friction-work quadrature comparison.

\section{Complete parameter-intervention results}
\label{app:allresults}
\begin{table*}[t]
\caption{Frozen \stater{} component on the new test set. Each row contains 20 trajectories. Torque and position entries are means of trajectory-level RMS values; energy is the mean of each trajectory's maximum cumulative residual. Position uses the maximum per-channel RMS. The torque standard deviation describes trajectories, not independent training seeds.}
\centering
\small
\begin{tabular}{lrrrr}
\toprule
Condition & Torque RMS mean $\pm$ std (N\,m) & Energy error (J) & Position error (mm) & Torque passes\\
\midrule
Nominal & $0.444\!\pm\!0.454$ & 3.397 & 61.6 & 10/20 \\
Viscous & $4.318\!\pm\!4.148$ & 121.155 & 109.6 & 0/20 \\
Coulomb & $0.442\!\pm\!0.448$ & 3.220 & 61.7 & 10/20 \\
Mixed & $4.274\!\pm\!4.188$ & 120.860 & 111.5 & 0/20 \\
Payload & $1.246\!\pm\!1.645$ & 1.955 & 59.9 & 9/20 \\
Density & $0.537\!\pm\!0.536$ & 3.087 & 55.1 & 10/20 \\
Size & $0.514\!\pm\!0.532$ & 10.248 & 54.8 & 10/20 \\
Slider & $1.663\!\pm\!2.375$ & 3.411 & 81.2 & 5/20 \\
\bottomrule
\end{tabular}
\end{table*}

All 160 video-conditioned outputs are finite and satisfy the differentiated constraints, yet fail closure and point prediction. Their velocities and accelerations therefore preserve an incorrect geometric level set. Because the raw coordinates never form a valid physical trajectory, native torque and energy residuals remain unavailable.

For the state component, all 160 trajectories are measured and satisfy all three constraint criteria. Motor-torque passes occur on 54/160 trajectories and conditional-prediction passes on 21/160. Their family-balanced rates are $30.0\%$ and $7.5\%$, respectively. No trajectory passes the energy criterion, hence none passes jointly.

The macro average assigns equal weight to four-bar and crank-slider families, then equal weight to the seven four-bar conditions. It therefore differs from the pooled 54/160 or 21/160 fractions. At half of every tolerance the joint rate remains zero; at twice every tolerance it is zero for the video interface and $3.57\%$ for the state control.

The continuous tolerance-ratio composite is $33.37/100$ for the video interface and $56.88/100$ for the state component. Almost all of the video score comes from differentiated constraints, which preserve a nonzero geometric level set. Raw residuals and strict joint rates distinguish this partial consistency from full mechanical qualification.

\section{Structural probes and parameter identification}
\label{app:probes}
A positive-semidefinite damping law supplies a dissipative force:
\begin{equation}
Q_{\rm diss}=-D_\theta(q,\dot q)\dot q,\qquad D_\theta\succeq0,
\label{eq:design-dissipation}
\end{equation}
which guarantees $\dot q\T Q_{\rm diss}\leq0$.

Under unsaturated proportional servo control, the velocity sensitivity of acceleration $\alpha_\pi$ combines intrinsic dynamics and actuation gain:
\begin{equation}
\frac{\partial\alpha_\pi}{\partial\omega}
=\frac{\partial\alpha}{\partial\omega}
-k_p\frac{\partial\alpha}{\partial\tau}.
\label{eq:controller-confound}
\end{equation}

Known geometry and inertia-to-mass ratios give a linear system for identifying masses and friction coefficients from calibrated motion and forcing:
\begin{equation}
\begin{aligned}
y&=Y(q,\dot q,\ddot q)\beta,\\
\beta&=(m_1,m_2,m_3,m_p,b_1,\ldots,b_4,f_{c1},\ldots,f_{c4})\T.
\end{aligned}
\label{eq:identification}
\end{equation}

\begin{samepage}
KeyCLD's Cartesian acceleration combines the learned force $f$ with a constraint reaction $\lambda$ that enforces the differentiated constraints:
\begin{align}
\widehat a&=M_x^{-1}(f-D\Phi\T\lambda),\\
\lambda&=(D\Phi M_x^{-1}D\Phi\T)^{-1}
\left[D\Phi M_x^{-1}f+D^2\Phi[\dot x,\dot x]\right].
\label{eq:keycld}
\end{align}
\end{samepage}
For the velocity-reversal probe we use angles evenly spaced from $-\pi$ to $\pi$, inclusive, and speed magnitudes $\{0.25,1,4,12\}$ rad/s. Thus the 73 angle samples include two representations of the periodic endpoint. There are 292 matched reversal pairs per friction condition. We hold the motor input at $1$ N\,m and evaluate each pair at the same physical pose.

The architecture's evenness follows from its position-only mass, potential, actuation, and constraint derivatives: changing $\dot x$ to $-\dot x$ leaves both the force and the quadratic term $D^2\Phi[\dot x,\dot x]$ unchanged, so it preserves the entire constrained acceleration. Smooth viscous and Coulomb terms instead change sign at fixed pose. The reversal probe therefore quantifies the dissipative response that this architecture cannot represent, extending the original KeyCLD discussion of absent dissipation~\cite{daems2024keycld} to the adapted mechanism.

For the scale probe, multiply the point-mass root parameters by $\sqrt{k}$ and the output layers of the potential and input networks by $k$. Then $M_x^{-1}$ scales by $1/k$, the constraint solve yields $\lambda' = k\lambda$, and the acceleration remains invariant. The recorded single-state probes at $k=0.25$ and $k=4$ both have zero numerical acceleration change. The algebraic statement applies to every regular state under these parameter transformations.

\begin{table}[t]
\caption{Behavioral effective-inertia errors from calibrated $\pm1$ N\,m action queries at zero speed. Error is the dimensionless fraction $\lvert\widehat{\mathcal I}_{\rm resp}/\mathcal I-1\rvert$, evaluated on positive gains.}
\label{app:torque-inertia}
\centering
\small
\begin{tabular}{lrrr}
\toprule
Condition & Positive & Median error & Mean error\\
\midrule
Nominal & 71/73 & 0.3516 & 3.4918 \\
Payload & 71/73 & 0.4156 & 2.6752 \\
Density & 71/73 & 0.5122 & 2.4735 \\
Size & 66/73 & 0.8542 & 5.6663 \\
Slider & 64/73 & 0.5185 & 3.6548 \\
\bottomrule
\end{tabular}
\end{table}

Effective inertia combines the learned mass and input-map responses. Its discrepancy measures their combined actuator calibration. The positive-gain counts specify the measured response set. The classical control below identifies physical masses and friction from calibrated mechanical equations.

For that control, the known geometry gives mass-linear bases for $M,c,G$, with prescribed inertia-to-mass ratios. We form one equation per sampled state,
\begin{equation}
y=Y(q,\dot q,\ddot q)\beta,
\end{equation}
where $\beta$ contains the crank, coupler, rocker/slider, and payload masses, four viscous coefficients, and four smooth Coulomb magnitudes. The known rotor and external-force terms are moved to $y$. We normalize regressor columns on the fitting split and solve nonnegative least squares. Seeds 1000--1011 fit and 1012--1019 test; no neural output is used.

\begin{table*}[t]
\caption{Exact-state physical parameter control on the extended dataset, with known geometry, inertia-to-mass ratios, and SI torque. Mass error is the largest absolute relative error among four masses, expressed as a fraction.}
\label{app:identification}
\centering
\small
\begin{tabular}{lrrrr}
\toprule
Condition & Rank & Normalized condition number & Maximum mass error & Test torque RMS (N\,m)\\
\midrule
Nominal & 12/12 & 483.3 & $1.28\!\times\!10^{-6}$ & $8.72\!\times\!10^{-8}$ \\
Viscous & 12/12 & 119.9 & $0.67\!\times\!10^{-6}$ & $7.03\!\times\!10^{-8}$ \\
Coulomb & 12/12 & 494.9 & $0.56\!\times\!10^{-6}$ & $7.64\!\times\!10^{-8}$ \\
Mixed & 12/12 & 123.4 & $1.26\!\times\!10^{-6}$ & $3.37\!\times\!10^{-8}$ \\
Payload & 12/12 & 300.1 & $0.44\!\times\!10^{-6}$ & $8.55\!\times\!10^{-8}$ \\
Density & 12/12 & 250.6 & $0.90\!\times\!10^{-6}$ & $8.54\!\times\!10^{-8}$ \\
Size & 12/12 & 1235.7 & $1.14\!\times\!10^{-6}$ & $9.41\!\times\!10^{-8}$ \\
Slider & 12/12 & 436.0 & $1.33\!\times\!10^{-6}$ & $9.74\!\times\!10^{-8}$ \\
\bottomrule
\end{tabular}
\end{table*}

The mixed-friction fit has maximum viscous coefficient error $5.85\times10^{-9}$ N\,m\,s/rad and maximum Coulomb torque error $4.75\times10^{-8}$ N\,m. Prescribed simulator parameters supply the SI reference.

\section{Reproducibility and study scope}
\label{app:reproducibility}
All models use fixed final-epoch checkpoints, with gradient updates and sequence exposures specifying the training budget across GPU and CPU environments. Released model modules and adaptation scripts define the predictors, physical readouts, and independent mechanism equations. The source package provides the figure assets, tables, and evidence mappings needed to reproduce the reported comparisons.

Forecasts use the observed prefix, public geometry and camera calibration, and commands. Exact-state controls replace visual coordinates with true training and prefix states. Future reference states, measured actuator forces, and independent mass, inertia, and friction parameters define the SI evaluator. Predictor and reader fitting uses training identities; validation and calibration select readout and processing settings before evaluation on held-out test assemblies. Reconstruction and reset-window diagnostics use the later observations specified in their protocols to measure local representation accuracy.

The torque-excitation study uses one KeyCLD seed, while each shared-assembly arm uses three seeds. Friction and payload interventions query the frozen KeyCLD model across changed physical conditions. The shared study evaluates transfer to unseen assemblies within three trained mechanism families, retaining each model's command interface and native objective.

The dataset consists of synthetic mechanism images and trajectories, with no human subjects or personal data.

\section{Additional residual distributions}
The following distributions and equation substitution retain the torque-excitation protocol. Their eight operating conditions, constant-torque inputs, and calibrated tolerances differ from the multi-assembly servo experiment.
\begin{figure*}[t]
  \centering
  \includegraphics[width=\textwidth]{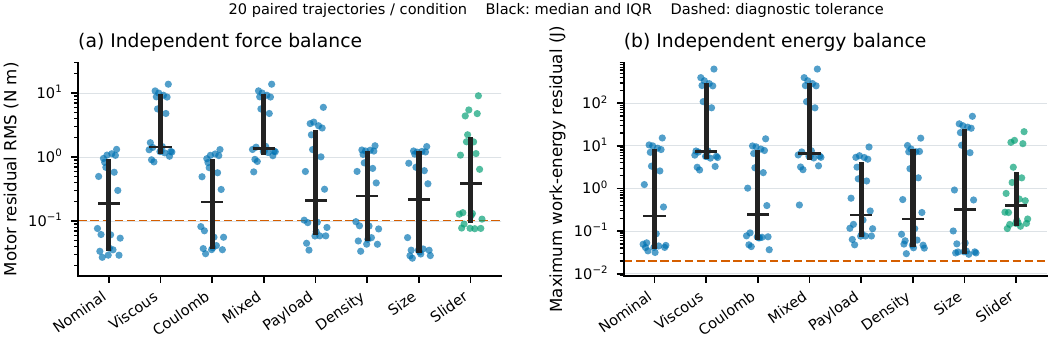}
  \caption{Independent physical errors of the frozen \stater{} component on the torque-excitation scoring set. Each condition contains 20 paired trajectories. Points are individual trajectories, black marks show the median and interquartile range, and dashed lines are the fixed diagnostic tolerances. All trajectories pass the three geometric/differential constraint criteria, while physical residuals vary substantially with condition. Both axes are logarithmic.}
  \label{fig:residuals}
\end{figure*}
\begin{figure}[t]
  \centering
  \includegraphics[width=\columnwidth]{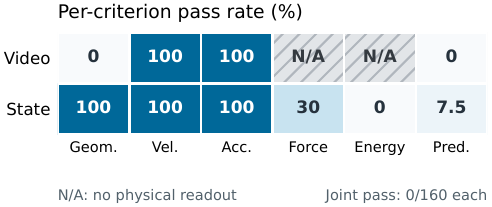}
  \caption{Torque-excitation pass profile, balanced across mechanism families and conditions. Hatched N/A cells have no valid physical readout and are distinct from measured zero-pass rates. All 160 state trajectories pass the three constraint criteria, while no trajectory passes work--energy balance.}
  \label{fig:passes}
\end{figure}
\begin{figure*}[t]
  \centering
  \includegraphics[width=\textwidth]{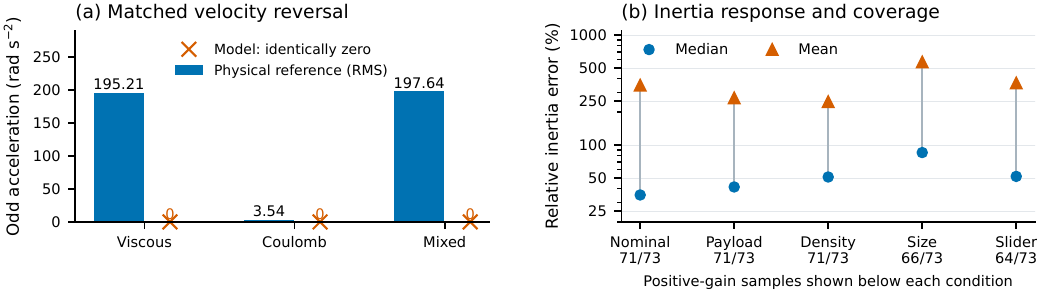}
  \caption{Constitutive probes of the frozen state component. (a) Matched velocity reversal gives nonzero reference odd acceleration and zero model response for all 292 pairs per condition. (b) Effective-inertia errors on positive input gains: both medians and means are shown on a log scale, with positive-gain counts below each condition. The large mean--median gaps retain the effect of near-zero gains.}
  \label{fig:probes}
\end{figure*}
\subsection{Interpreting a concrete equation mismatch}
\begin{table}[t]
\caption{Matched-state acceleration queries at the fixed first extended-test seed and $t=0.11$ s. Each condition uses its corresponding true pose and speed. These are local constitutive probes, separate from trajectory scoring.}
\label{tab:local}
\centering
\small
\begin{tabular}{lrrr}
\toprule
Condition & True $\alpha$ & Model $\alpha$ & $r_\tau$\\
& (rad/s$^2$) & (rad/s$^2$) & (N\,m)\\
\midrule
Nominal & 590.32 & 613.06 & 0.08196\\
Mixed friction & 48.46 & 141.58 & 0.53881\\
Slider & $-144.72$ & 95.43 & 0.96611\\
\bottomrule
\end{tabular}
\end{table}
The residual has a direct interpretation in the governing mechanical equation. In the fixed nominal example of \cref{tab:local}, the independent equation at the measured crank state is
\begin{equation}
\begin{split}
0.00360379\,\alpha
-0.00164571(25.51406)^2\\
-0.41765765=0.61435837+0.02407613.
\end{split}
\label{eq:worked}
\end{equation}
The terms are effective inertia, velocity-quadratic torque, gravity, motor torque, and external-force torque. It requires $\alpha=590.32$ rad/s$^2$, whereas the model queried at that same state predicts $613.06$ rad/s$^2$. Their difference corresponds to $0.08196$ N\,m. In the slider query the predicted acceleration has the opposite sign to the physical reference. These examples illustrate the units and meaning of the test; the overall conclusions use all paired trajectories. A locally small force discrepancy can also coexist with an unacceptable accumulated energy error over the full rollout.

\section{Assembly partitions and observation preprocessing}
\label{app:corpus-data}
The multi-assembly study uses a frozen subset of the accepted planar-linkage corpus. Each assembly has its own geometry, rigid-body properties, and three operating trajectories. Table~\ref{app:corpus-splits} counts distinct assemblies. Each trajectory has multiple rendered views, but the current experiment uses the front view and counts each operating trajectory once. Video availability determines the frozen training subset before model evaluation.

\begin{table}[t]
\caption{Multi-assembly partition. Each assembly contributes three operating conditions. Validation and numerical calibration occupy separate subsets of the original validation partition.}
\label{app:corpus-splits}
\centering\small
\begin{tabular}{lrrrr}
\toprule
Family & Train & Val. & Cal. & Test\\
\midrule
Crank-rocker & 96 & 6 & 6 & 12\\
Crank-slider & 92 & 6 & 6 & 12\\
Double-crank & 83 & 6 & 6 & 12\\
\midrule
Assemblies & 271 & 18 & 18 & 36\\
Trajectories & 813 & 54 & 54 & 108\\
\bottomrule
\end{tabular}
\end{table}

Physical parameters include all moving hardware in the body mass, center of mass, and inertia. Independent rigid-body equations check the simulated mass and bias terms. MuJoCo~\cite{todorov2012mujoco} supplies the trajectories; Blender replays their body transforms for rendering. The physics step is $1/4800$ s, reference recording is $240$ Hz, and videos have 120 frames at $30$ Hz. Finite numerical constraint compliance is retained in the references and measured by calibration.

Source videos have $640\times640$ pixels. Orthographic calibration rectifies the motion plane at $Y=0$ into $64\times64$ and $256\times256$ RGB, with a fixed world half extent $0.65$ m and center $(0.20,0.35)$ m. Physical coordinates satisfy
\begin{equation}
x_{\rm world}=0.65x_{\rm normalized}+(0.20,0.35).
\end{equation}
Plate thickness and its parallax remain in the observations. Source camera framing used the reference trajectory extents; cameras then remain fixed. The evaluation uses these recorded cameras and their given calibration.

Training uses frames 30--79. Evaluation observes frames 30--32, initializes the ODE at the central frame, and predicts through frame 119. The native grid contains 705 states from $1+8/240$ to $1+712/240$ s. The 689 samples from $1.10$ s onward are scored. Energy is anchored at the last observed time, $1+16/240$ s.

Velocity commands have settled by $t=1$ s. Crank-rockers and sliders use proportional gain $2$ N\,m\,s/rad and a $5$ N\,m torque bound. Double-cranks use gain $12$ N\,m\,s/rad and a $12$ N\,m bound. KeyCLD's wrapper recomputes torque from its predicted crank velocity, with known external force providing two additional action channels. CON learns the closed-loop evolution from command and controller metadata. The evaluator recomputes physical servo torque for both models.

The three commands are $1$, $3$, and $5$ rad/s; the last condition jointly increases speed and multiplies payload mass by $1.75$. All three occur in training. The test set therefore measures generalization to new assemblies within these families and operating conditions.

\section{KeyCLD training and shared numerical calibration}
The architecture and video objective use the same released modules as the torque-excitation experiment. Corpus training has 813 sequences, 40 epochs, and 32,520 updates per run. Seeds 42, 43, and 44 are used independently for the 64-pixel video arm and two state-supervised controls. For each seed, the three arms start from the same saved initial parameters. State supervision replaces both encoded training coordinates/velocities and the forecast initialization. State--Euler and State--RK4 therefore differ in training integration, while sharing their data and four-step loss windows. Visual parameters are unchanged in the state controls.

KeyCLD uses float32 computation in JAX 0.11.2. Adaptive validation rollouts evaluate the fixed final-epoch checkpoints. Paired Euler/RK4 controls receive equal updates and sequence exposures.

The registered physical thresholds are $1$ mm, $0.01$ m/s, $0.1$ m/s$^2$, $0.25$ N\,m, $0.05$ J, and $1$ cm for the six criteria. The last three quantities are torque RMS, maximum prefix energy error, and maximum-channel point RMS. The corpus torque/energy thresholds use a common $5$ N\,m reference scale. Thus $0.25$ N\,m is $5\%$ of the crank-rocker/slider bound and $2.08\%$ of the double-crank bound.

Independent calibration uses 54 trajectories from 18 separate assemblies. All positive controls pass. Maximum geometric error is $7.23\times10^{-6}$ m, velocity-constraint RMS $3.84\times10^{-5}$ m/s, acceleration-constraint RMS $6.97\times10^{-4}$ m/s$^2$, torque RMS $2.06\times10^{-5}$ N\,m, and energy error $2.23\times10^{-4}$ J. Geometrically admissible persistence and $0.8$-time-scaling controls each pass 0/54 jointly.

Raw predictions and native derivatives define the native evaluation track. Complete force/energy measurement requires valid geometry and a regular physical-coordinate readout throughout the scored interval; unavailable terms remain missing. Native integration uses the same adaptive solver with a 4,096-step interval cap. A 10 s integration wall-time budget and a separate 120 s initialization/compilation budget retain timeouts as explicit failed cases.

\section{KeyCLD 64-pixel and state-control results}
\label{app:corpus-results}
Table~\ref{app:corpus-seeds} reports finite outputs and physical readout coverage on the same 108 test trajectories per seed. The 324 predictions retain pairing across 36 independent assemblies. The video seed-44 run contains 13 test timeouts. All 648 test predictions from the two state-supervised arms are finite and physically measurable.

\begin{table*}[t]
\caption{All multi-assembly test runs. Every checkpoint completes 40 epochs and 32,520 updates. Finite output and a complete physical readout are distinct requirements.}
\label{app:corpus-seeds}
\centering\small
\begin{tabular}{lrrrrrr}
\toprule
Arm & Seed & Finite & Timeouts & Physical readout & $S_{\rm comp}$ & Joint (\%)\\
\midrule
Video & 42 & 108/108 & 0 & 0/108 & 14.78 & 0 \\
Video & 43 & 108/108 & 0 & 0/108 & 21.14 & 0 \\
Video & 44 & 95/108 & 13 & 0/108 & 11.28 & 0 \\
State--Euler & 42 & 108/108 & 0 & 108/108 & 52.66 & 0 \\
State--Euler & 43 & 108/108 & 0 & 108/108 & 52.62 & 0 \\
State--Euler & 44 & 108/108 & 0 & 108/108 & 52.66 & 0 \\
State--RK4 & 42 & 108/108 & 0 & 108/108 & 64.36 & 0 \\
State--RK4 & 43 & 108/108 & 0 & 108/108 & 64.22 & 0 \\
State--RK4 & 44 & 108/108 & 0 & 108/108 & 64.32 & 0 \\
\bottomrule
\end{tabular}
\end{table*}

\begin{table*}[t]
\caption{Mechanism-level state-control results. Each row averages 108 predictions: 12 assemblies, three conditions, and three seeds. Errors are means of trajectory statistics. Point error is maximum-channel RMS; energy error is maximum absolute prefix imbalance. All rows have 100\% passes on the three constraint criteria.}
\label{app:corpus-families}
\centering\small
\begin{tabular}{llrrrrrr}
\toprule
Arm & Family & Point (mm) & Torque (N\,m) & Energy (J) & Point pass (\%) & Torque pass (\%) & Energy pass (\%)\\
\midrule
State--Euler & Crank-rocker & 69.20 & 1.907 & 9.535 & 25.00 & 25.00 & 0 \\
State--Euler & Crank-slider & 81.31 & 3.172 & 38.040 & 0.00 & 0.00 & 0 \\
State--Euler & Double-crank & 266.68 & 8.612 & 19.785 & 0.00 & 0.00 & 0 \\
State--RK4 & Crank-rocker & 12.91 & 0.320 & 1.544 & 42.59 & 17.59 & 0 \\
State--RK4 & Crank-slider & 32.48 & 1.434 & 6.228 & 35.19 & 0.00 & 0 \\
State--RK4 & Double-crank & 3.71 & 0.391 & 1.004 & 97.22 & 4.63 & 0 \\
\bottomrule
\end{tabular}
\end{table*}

The paired RK4-minus-Euler differences are $-122.69$ mm for point error (95\% interval $[-128.17,-115.21]$), $-3.849$ N\,m for torque RMS ($[-4.059,-3.578]$), and $-19.528$ J for energy error ($[-21.744,-17.373]$). Bootstrap draws resample three training seeds and 12 assembly identities within each family, using the same draw in both arms and retaining all conditions together. We use 2,000 draws with the registered resampling seed. Family and condition averages receive equal weight.

Table~\ref{app:state-components} decomposes the paired continuous-score change. Both arms have 100\% kinematic pass rates; torque passes are 8.33\%/7.41\% and energy passes are zero. The continuous torque score rises from 6.98 to 29.93, reflecting residual magnitudes within the pass and fail groups. Tracking contributes 7.81 composite points (67.02\% of the gain), torque contributes 3.82 (32.83\%), and energy contributes 0.0175 (0.15\%). The total paired gain is 11.65 points with 95\% interval $[10.87,12.48]$, computed with the same seed/assembly protocol (2,000 draws, resampling seed 20261016). Scores are transformed per trajectory before averaging; transforming a mean residual gives a different quantity.

\begin{table*}[t]
\caption{Exact-state KeyCLD continuous component scores. G/V/A are closure/velocity/acceleration, F/E are torque/energy, and P is tracking. Each component uses $100/(1+\rho^2)$; $S_{\rm comp}$ is their arithmetic mean. The final row gives each component's contribution to the RK4-minus-Euler composite gain. Both arms have 324/324 coverage and zero joint passes.}
\label{app:state-components}
\centering\small
\begin{tabular}{lrrrrrrr}
\toprule
Training & G & V & A & F & E & P & $S_{\rm comp}$\\
\midrule
State Euler & 100.00 & 100.00 & 100.00 & 6.98 & 0.48 & 8.44 & 52.65 \\
State RK4 & 100.00 & 100.00 & 100.00 & 29.93 & 0.58 & 55.29 & 64.30 \\
\midrule
$\Delta/6$ (points) & 0.0002 & 0.0000 & 0.0000 & 3.82 & 0.02 & 7.81 & 11.65 \\
\bottomrule
\end{tabular}
\end{table*}

Joint pass counts are zero in every observed group. Their empirical bootstrap distributions are also zero; Tables~\ref{app:corpus-families} and~\ref{app:state-components} report the paired residual and continuous-score differences.

\section{KeyCLD response-query grids and inertia fitting}
\label{app:corpus-probes}
A usable response gain must be finite and exceed $10^{-8}$ rad\,s$^{-2}$/(N\,m). Failed checks include negative, near-zero, and nonfinite gains. The main figure reports this fraction alongside the inertia distribution of usable responses.

These post-hoc diagnostics query frozen state-supervised checkpoints. Six predetermined reference states from each of 108 test trajectories give 648 states per checkpoint. The probe supplies true physical coordinates and varies the applied motor torque independently of the velocity servo, isolating response to controlled mechanical inputs.

At zero speed, we query $\tau=\pm0.25$ N\,m with the known external force fixed. The finite difference gives $\partial\widehat\alpha/\partial\tau$ and its reciprocal defines behavioral effective inertia for usable gains. Tables~\ref{app:torque-inertia} and~\ref{app:corpus-response} express absolute relative errors as dimensionless fractions. Figure~\ref{fig:constitutive}b instead plots the inertia ratio $\widehat{\mathcal I}_{\rm resp}/\mathcal I$; its pooled medians are $74.9\times$ for Euler and $117.3\times$ for RK4. The learned input matrix is unconstrained, so response calibration combines actuation and inertia.

\begin{table*}[t]
\caption{Counterfactual SI response of the corpus state controls. Inertia error is $\lvert\widehat{\mathcal I}_{\rm resp}/\mathcal I-1\rvert$, conditioned on positive gain. Friction error is the RMS equivalent torque of the missing velocity-odd response. Each checkpoint uses the same 648 physical states.}
\label{app:corpus-response}
\centering\small
\begin{tabular}{lrrrrr}
\toprule
Arm & Seed & Positive gains & Median inertia error & Mean inertia error & Friction RMS (N\,m)\\
\midrule
State--Euler & 42 & 445/648 & 74.24 & 220.28 & 0.039 \\
State--Euler & 43 & 458/648 & 73.68 & 325.58 & 0.039 \\
State--Euler & 44 & 443/648 & 73.90 & 243.99 & 0.039 \\
State--RK4 & 42 & 570/648 & 115.39 & 210.19 & 0.039 \\
State--RK4 & 43 & 570/648 & 113.45 & 206.87 & 0.039 \\
State--RK4 & 44 & 575/648 & 120.12 & 224.87 & 0.039 \\
\bottomrule
\end{tabular}
\end{table*}

For friction, independent crank speeds are matched at $\pm1$ rad/s with motor torque fixed at zero. The learned odd acceleration is exactly zero for all six checkpoints; the reference equivalent friction-torque RMS is $0.03885$ N\,m. Holding physical torque fixed removes the public servo's known damping from this constitutive test.

A separate kinetic representation control uses all 360 physically specified planar assemblies, including the 17 without accepted videos in the training-data snapshot. For each of 1,080 conditions, 41 recorded poses form the positive point-mass fitting problem from the main paper's reduced-inertia equation. Separate per-condition fits achieve relative RMS error below $1.55\times10^{-15}$ over these recorded poses. Fitting one shared mass vector across all conditions gives $22.55\%$ relative RMS discrepancy. These evaluator fits quantify the kinetic representation over the specified reference states.

The per-condition and shared fits separate local representational capacity from consistency across heterogeneous mechanisms. Forecast accuracy also depends on learned potential and actuation, whose identification is affected by the state-dependent servo input. Checking excitation and regressor rank under the known controller therefore complements the full-rank independent-torque control.

\section{CON configuration and shared-data extension}
\label{app:con}
The CON adaptation uses the released \texttt{node-con-iae} implementation~\cite{stolzle2024con}, revision \texttt{cef5441f0201ed28e3ac9e1d3e7d4de7d52c4a30}. Its CNN $\beta$-VAE and coupled-oscillator dynamics retain the upstream mathematical implementation. At 64 pixels, the visual model has 67,271,267 parameters and eight latent coordinates; the separate state component has 13,360 parameters. A five-layer, width-30 input autoencoder maps the public command and geometry information to this latent dynamical system.

Both models use the same assembly manifest, source observations, three-frame prefix, calibration trajectories, and forecast grid. Each seed evaluates 108 test sequences from 36 held-out identities. Table~\ref{app:two-model-budget} records the native training budgets. CON includes the final batch of 13 sequences, giving 51 updates per epoch. KeyCLD evaluates 43 overlapping four-step windows per sequence, while CON integrates one 48-interval rollout.

\begin{table*}[t]
\caption{64-pixel video and state-control training configurations. Counts are per run. All arms use 813 training sequences and seeds 42, 43, and 44. A sequence use is one sequence in one epoch; temporal loss windows differ by model.}
\label{app:two-model-budget}
\centering\small
\begin{tabular}{llrrrrl}
\toprule
Model & Arms & Epochs & Batch & Updates & Sequence uses & Training integration\\
\midrule
KeyCLD & Video, State--Euler, State--RK4 & 40 & 1 & 32,520 & 32,520 & Euler / RK4 substeps\\
CON & Video, State--Dopri5 & 50 & 16 & 2,550 & 40,650 & Dopri5, $0.01$ s steps\\
\bottomrule
\end{tabular}
\end{table*}

\paragraph{Command and geometry input.}
The fixed 14-dimensional input contains command speed, servo gain, torque limit, in-plane force $(F_x,F_z)$, and nine geometry entries $(A_x,A_z,D_x,D_z,l_1,l_2,l_3,h_s,\mathrm{slider})$. Their normalization divisors are $(5,12,12,1,1)$ followed by nine ones. CON learns command-conditioned closed-loop evolution, and its input decoder reconstructs the normalized metadata. The released model's positive-definite inverse inertia, stiffness, and damping matrices act in latent coordinates. We evaluate this interface through physical readouts and trajectories; the independent evaluator computes SI motor torque from the public servo law.

\paragraph{Objective and optimization.}
CON Video retains latent rollout MSE, static RGB reconstruction, dynamic RGB reconstruction, KL regularization, and input reconstruction. Their weights are $0.3733687489$, $1$, $83.7248326772$, $0.0002006838$, and $10$. AdamW uses peak learning rate $0.0184869909$, weight decay $0.0000553401$, five warmup epochs, and cosine decay thereafter. The training objective integrates the full 50-frame window with Dopri5 and fixed $0.01$ s steps. State trains the same dynamical component and input autoencoder against true normalized $B,C,P$ coordinates and velocities, padded with two zero coordinates. Its visual losses are inactive. State-training integrations use float64 ODE states, with predicted coordinates cast to float32 for the loss. Section~\ref{app:con-integrators} adds a paired state-training integrator control.

\paragraph{Initialization and physical readout.}
For Video, the encoder's Jacobian--vector product maps the centered difference of the three observed images into latent velocity at the middle frame. State starts from exact middle-frame coordinates and velocity. Frozen rollouts integrate float64 ODE states using adaptive Dopri5 with relative/absolute tolerances $10^{-7}/10^{-9}$ and native derivatives on the common 240 Hz grid. RGB decoding uses float32 latent inputs.

After video training, a single affine latent-to-point map is fitted on 40,650 training-frame states (813 sequences, 50 frames each). It provides a fixed coordinate readout for validation and testing. Validation RMSE averages squared error over the six coordinates and observed frames before averaging trajectories. Future point error instead uses the shared maximum-channel trajectory RMS. This privileged readout gives an interpretable diagnostic interface, distinct from KeyCLD's validation-selected keypoint permutation. State reads the first six latent coordinates directly. Both native readouts produce unconstrained coordinates; Section~\ref{app:geometry-readout} defines the geometric intervention.

\section{CON 64-pixel and state-control results}
\begin{table*}[t]
\caption{Native criterion pass rates (\%) and continuous $S_{\rm comp}$ with seed/assembly bootstrap intervals for the five original assembly arms. Physical counts complete SI measurements. Table~\ref{app:state-components} separately gives continuous state-control components.}
\label{app:native-passes}
\centering\small
\begin{tabular}{lrrrrrrrrl}
\toprule
Interface & Physical & Geom. & Vel. & Acc. & Torque & Energy & Pred. & Joint & $S_{\rm comp}$ [95\% CI]\\
\midrule
\multicolumn{10}{l}{\textit{KeyCLD}}\\
Video & 0/324 & 0.00 & 22.22 & 65.12 & N/A & N/A & 0.00 & 0 & 15.73 [11.36, 20.97] \\
State--Euler & 324/324 & 100.00 & 100.00 & 100.00 & 8.33 & 0.00 & 8.33 & 0 & 52.65 [52.30, 52.96] \\
State--RK4 & 324/324 & 100.00 & 100.00 & 100.00 & 7.41 & 0.00 & 58.33 & 0 & 64.30 [63.60, 65.03] \\
\midrule
\multicolumn{10}{l}{\textit{CON}}\\
Video + affine & 0/324 & 0.00 & 0.93 & 33.02 & N/A & N/A & 0.00 & 0 & 7.23 [3.72, 9.73] \\
State & 0/324 & 0.00 & 0.00 & 3.40 & N/A & N/A & 0.00 & 0 & 1.91 [1.22, 2.71] \\
\bottomrule
\end{tabular}
\end{table*}

\label{app:con-results}
The six native Dopri5 runs each complete 50 epochs and 2,550 updates, with all 108 test and 54 validation forecasts finite. Table~\ref{app:con-seeds} reports every seed, and the common SI evaluator measures all 648 CON test predictions against the shared engineering targets.

\begin{table*}[t]
\caption{CON seed results. Finite and complete physical coverage refer to 108 test trajectories. Point error is mean maximum-channel future RMS. RGB MSE uses future frames 33--119 and is displayed in units of $10^{-4}$. Reader validation uses observed validation images and coordinate RMSE, not future forecast error. State has no RGB decoder or fitted reader.}
\label{app:con-seeds}
\centering\small
\begin{tabular}{lrrrrrrr}
\toprule
Arm & Seed & Finite & Physical & $S_{\rm comp}$ & Point (mm) & RGB MSE & Reader val. (mm)\\
\midrule
Video + affine & 42 & 108/108 & 0/108 & 9.51 & 123.84 & 3.1102 & 90.39 \\
Video + affine & 43 & 108/108 & 0/108 & 3.53 & 124.75 & 3.1278 & 86.42 \\
Video + affine & 44 & 108/108 & 0/108 & 8.64 & 121.86 & 3.0027 & 93.54 \\
State & 42 & 108/108 & 0/108 & 1.18 & 167.48 & N/A & N/A \\
State & 43 & 108/108 & 0/108 & 2.75 & 128.01 & N/A & N/A \\
State & 44 & 108/108 & 0/108 & 1.80 & 131.68 & N/A & N/A \\
\bottomrule
\end{tabular}
\end{table*}

Video's RGB MSE averages $3.0802447\times10^{-4}$ across seeds, families, and conditions. Last-observed-image persistence gives $4.9398983\times10^{-4}$ on the same future targets, a $37.6456\%$ relative reduction. Decoded RGB is stored as clipped 8-bit pixels; squared errors use $[0,1]$ normalization. This pixel comparison concerns the assembly corpus. KeyCLD's torque-excitation RGB and VPT diagnostics use different data and are reported separately.

The video reader's validation RMSE spans $86.42$--$93.54$ mm. Across future test trajectories, Video and State have mean maximum-channel point RMS of $123.48$ and $142.39$ mm, respectively. Their mean trajectory-maximum geometry errors are $137.67$ and $170.40$ mm. No trajectory in either arm passes geometry, yielding zero complete physical coverage and zero joint qualification. Native torque and energy values are unavailable because the physical-coordinate interface fails geometric validity. Continuous $S_{\rm comp}$ is $7.23$ with 95\% interval $[3.72,9.73]$ for Video and $1.91$ with $[1.22,2.71]$ for State. Intervals use 2,000 seed/assembly resamples with conditions kept together.

The extension tests sustained servo-driven closed-chain motion, a periodic regime discussed in CON's original formulation~\cite{stolzle2024con}. Its state control supervises padded Cartesian coordinates directly. Maintaining the curved closure manifold then depends on the learned oscillator flow, and the observed forecasts depart from it despite exact-state supervision. Together with KeyCLD's admissible state trajectories and residual force errors, these results locate complementary failures under the same physical criteria.

\section{Matched 64-pixel video and inference diagnostics}
\label{app:matched-video}
Frozen-checkpoint diagnostics compare both models on the same 108 assembly sequences for each of three seeds. Framewise reconstruction encodes each true image; forecasting evolves from the three-frame prefix. Reconstruction and reset-window comparisons separate local representation error from open-loop prediction.

\begin{table*}[t]
\caption{Matched assembly-video diagnostics. RGB MSE is in units of $10^{-4}$ on future frames 33--119. Four-step reset error measures self-consistency against later observed encodings, in millimeters through each model's distinct fixed point readout. Means balance seed, family, and condition.}
\label{app:matched-video-table}
\centering\small
\begin{tabular}{lrrrrrrr}
\toprule
Model & Reconstruction & Forecast & Finite forecast & Persistence & Reset Euler & Reset RK4 & Reset hold\\
\midrule
KeyCLD & 2.7100 & 4.9595 & 311/324 & 4.9399 & 5.208 & 5.186 & 5.272 \\
CON & 2.7952 & 3.0802 & 324/324 & 4.9399 & 14.416 & 14.341 & 17.605 \\
\bottomrule
\end{tabular}
\end{table*}

Each 90-frame test sequence supplies 84 local windows centered at observed frames 1--84 in the cropped sequence. Encodings initialize positions, while KeyCLD's centered differences and tangent projection or CON's encoder Jacobian--vector product initialize velocities. Each window predicts four subsequent encodings using either one Euler step or four RK4 substeps per video interval. The frozen encoder, dynamics, and readout are also compared with training-set mean-image and mean-pose controls, using the same temporal-variance motion mask.

KeyCLD's assembly forecast MSE is about $1.83$ times its framewise reconstruction error and is close to persistence. On the 311 common finite cases, KeyCLD/CON forecast MSE is $(4.9595,3.0914)\times10^{-4}$. The paired CON-minus-KeyCLD difference is $-1.8680\times10^{-4}$, with seed/assembly bootstrap interval $[-2.4067,-1.1620]\times10^{-4}$.

\begin{figure}[t]
\centering
\includegraphics[width=\linewidth]{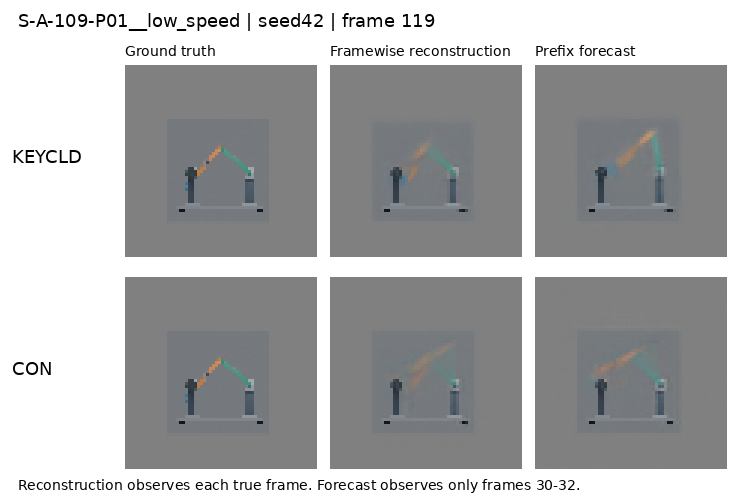}
\caption{Last forecast frame (119) of the first frozen test case, seed 42. Reconstruction uses the current true image to test visual representation, while forecasting uses frames 30--32 to test temporal prediction.}
\label{app:matched-video-figure}
\end{figure}

\paragraph{Fixed-weight inference solvers.}
For each video checkpoint, Euler and RK4 use 1, 4, and 16 substeps per $1/30$ s interval. The 3,888 trajectories share saved native initial states and retain each model's parameters and precision. For native timeouts without a finite saved initial state, the original three-frame initializer is rerun. The six solver settings therefore compare fixed-weight evolution from common observed prefixes.

\begin{table*}[!t]
\caption{Frozen video inference sweeps. RGB MSE is in units of $10^{-4}$. Native difference is point-coordinate RMS in millimeters; parentheses count common finite cases. Finite trajectories have denominator 324. KeyCLD's one-step Euler value, $6.389\times10^{15}$ mm, records numerical divergence; refinement to four/sixteen steps gives 1.25/0.347 mm.}
\label{app:inference-solvers}
\centering\small
\begin{tabular}{llrrr}
\toprule
Model & Solver / steps & Finite & RGB MSE & Native difference\\
\midrule
KEYCLD & euler / 1 & 311/324 & 5.17041 & 6.389e+15 (299) \\
KEYCLD & euler / 4 & 322/324 & 4.98690 & 1.25 (310) \\
KEYCLD & euler / 16 & 323/324 & 4.96358 & 0.3465 (311) \\
KEYCLD & rk4 / 1 & 321/324 & 4.97731 & 7.02e+11 (309) \\
KEYCLD & rk4 / 4 & 324/324 & 4.95350 & 0.02659 (311) \\
KEYCLD & rk4 / 16 & 324/324 & 4.95351 & 0.003339 (311) \\
CON & euler / 1 & 324/324 & 3.08396 & 1.731 (324) \\
CON & euler / 4 & 324/324 & 3.08098 & 0.4099 (324) \\
CON & euler / 16 & 324/324 & 3.08042 & 0.1011 (324) \\
CON & rk4 / 1 & 324/324 & 3.08024 & 5.512e-05 (324) \\
CON & rk4 / 4 & 324/324 & 3.08024 & 1.21e-05 (324) \\
CON & rk4 / 16 & 324/324 & 3.08024 & 1.209e-05 (324) \\
\bottomrule
\end{tabular}
\end{table*}

CON's refined solvers converge to nearly identical RGB error. KeyCLD's coarse Euler and RK4 predictions can contain extremely large finite coordinates even when decoded RGB error remains bounded. RK4 with four or sixteen substeps produces 324 finite trajectories in this fixed-weight sweep, compared with 311 in native inference. This isolates inference integration from the training-integrator intervention below.

\section{Paired CON training-integrator controls}
\label{app:con-integrators}

\begin{table*}[!t]
\caption{Additional CON state-training controls, averaged over three seeds. Point error uses maximum-channel future RMS; geometry uses the trajectory maximum, both in millimeters. The original State--Dopri5 arm remains a separate reference.}
\label{app:con-integrator-table}
\centering\small
\begin{tabular}{lrrr}
\toprule
Training solver & Point & Geometry & Raw physical\\
\midrule
EULER & 124.475 & 126.306 & 0/324 \\
RK4 & 137.382 & 142.066 & 0/324 \\
\bottomrule
\end{tabular}
\end{table*}
Six additional privileged-state runs compare Euler and RK4 across seeds 42--44. Within each seed, initial parameters and all 50 shuffled epoch orders match. Architecture, normalized state/action inputs, loss weights, optimizer schedule, batch size 16, and final epoch 50 are fixed. Both train on all 813 sequences per epoch, giving 2,550 updates and 40,650 sequence exposures. The state objective predicts 48 frame intervals from the middle observed training frame; Euler uses one step and RK4 four substeps per interval. Both use float64 ODE states and float32 coordinate losses, as in the native state control.

The six runs produce 972 native adaptive-Dopri5 forecasts: 648 test and 324 validation predictions. Table~\ref{app:con-integrator-table} reports the native coordinates. RK4 increases point error by $12.907$ mm, paired 95\% interval $[6.734,23.497]$, across all 324 test pairs. Both arms retain zero complete native physical coverage. Higher-order training integration thus increases trajectory error within this oscillator pipeline, alongside the reductions observed for KeyCLD.

\section{Common geometry-conditioned force and energy readout}
\label{app:geometry-readout}

\begin{table*}[!t]
\caption{Explicit geometric-readout intervention on seven assembly arms, $n=324$ per arm. Coverage counts complete force/energy trajectories. Correction is per-trajectory coordinate RMS over the saved grid; point error is maximum-channel future RMS. Correction and residual means use available cases with equal seed/family/condition weighting.}
\label{app:geometry-readout-table}
\centering\small
\begin{tabular}{lrrrrrr}
\toprule
Model / training & Raw physical & Readout physical & Correction (mm) & Torque (N\,m) & Energy (J) & Point (mm)\\
\midrule
KeyCLD Video & 0/324 & 311/324 & 120.4944 & 6.3830 & 0.9213 & 194.798 \\
KeyCLD State--Euler & 324/324 & 324/324 & 0.0009 & 4.5635 & 22.4534 & 139.062 \\
KeyCLD State--RK4 & 324/324 & 324/324 & 0.0006 & 0.7148 & 2.9253 & 16.368 \\
CON Video & 0/324 & 308/324 & 73.1848 & 6.8593 & 17.9601 & 154.833 \\
CON State--Dopri5 & 0/324 & 323/324 & 67.6754 & 6.5488 & 23.2825 & 132.278 \\
CON State--Euler & 0/324 & 317/324 & 54.7288 & 6.0279 & 11.8472 & 133.064 \\
CON State--RK4 & 0/324 & 322/324 & 58.0090 & 6.5413 & 20.4764 & 132.858 \\
\bottomrule
\end{tabular}
\end{table*}
The same post-hoc geometric intervention maps both architectures' frozen point predictions to admissible trajectories. It operates after rollout and defines an additional evaluation track, reported alongside native coverage and residuals.

Let $b=B-A$ be the predicted crank direction, with predicted derivatives $v_B,a_B$. We decode
\begin{align}
\theta&=\operatorname{atan2}(b_z,b_x),\qquad
\omega=\frac{b\times v_B}{\|b\|^2},\\
\alpha&=\frac{b\times a_B}{\|b\|^2}
-2\frac{b\T v_B}{\|b\|^2}\omega.
\end{align}
Public mechanism geometry determines $\widetilde q=\operatorname{IK}_{\mathcal G}(\theta)$ on the declared assembly branch. For angular-coordinate closure $\psi(q)=0$, write $J_q=[J_{q,0}\;J_{q,d}]$ and
\begin{align}
\gamma&=\begin{bmatrix}1\\-J_{q,d}^{-1}J_{q,0}\end{bmatrix},&
\gamma'&=\begin{bmatrix}0\\-J_{q,d}^{-1}D^2\psi[\gamma,\gamma]\end{bmatrix},\\
\dot{\widetilde q}&=\gamma\omega,&
\ddot{\widetilde q}&=\gamma\alpha+\gamma'\omega^2.
\end{align}
Forward kinematics maps predicted $B,v_B,a_B$ and public geometry to consistent points and analytic derivatives. Cases with predicted crank radius below $1$ mm or dependent-coordinate condition number above $10^{10}$ remain unavailable.

\paragraph{Calibration and coverage.}
All 54 independent calibration references pass after this map; persistence and $0.8$ time scaling each pass 0/54. Maximum calibration correction RMS is $2.481\,\mu$m, maximum torque RMS $2.17\times10^{-4}\,\Nm$, and maximum energy defect $1.21\times10^{-4}$ J, using the registered engineering targets.

The readout enables force and energy measurement for 308/324 CON Video and 323/324 native CON State trajectories, with average corrections of $73.185$ and $67.675$ mm. KeyCLD's state trajectories require less than $1\,\mu$m mean correction, while its video predictions require $120.494$ mm. The correction cost quantifies how far the specified reconstruction moves each predicted trajectory to obtain admissible coordinates.

Across the 315 CON Euler/RK4 pairs with both readouts available, RK4-minus-Euler torque difference is $0.560\,\Nm$ with interval $[-0.158,2.070]$, and energy difference is $9.084$ J with interval $[-0.370,27.863]$. Both intervals span zero: among these measurable trajectories, the force/energy effect remains unresolved, while the native point error increases across all 324 pairs.

\begin{figure*}[!t]
\centering
\includegraphics[width=.78\textwidth]{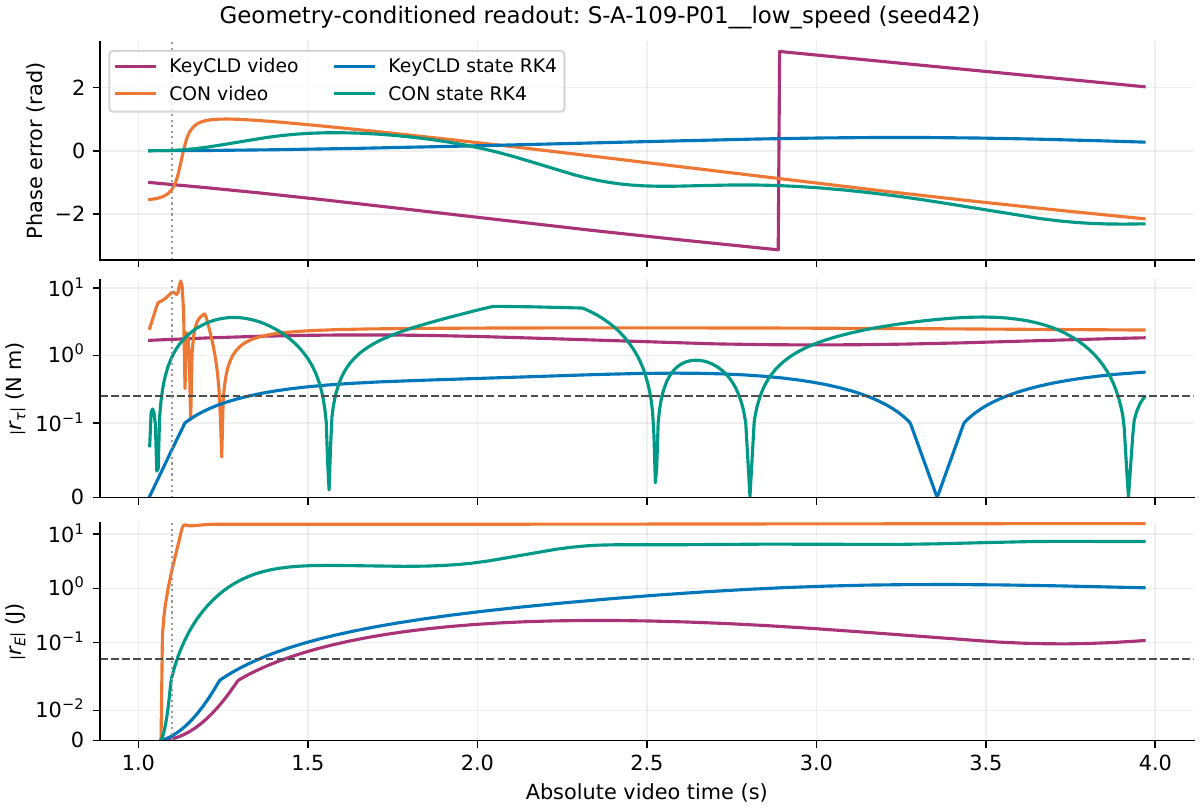}
\caption{Independent force/energy evaluation after the common geometric readout, using the first frozen test case and seed 42; video arms use 64-pixel inputs. Panels show wrapped phase error in $[-\pi,\pi]$, $\lvert r_\tau\rvert$, and $\lvert r_E\rvert$. Dashed lines show the $0.25\,\Nm$ and $0.05$ J criteria; the dotted line marks the first future frame. Work is anchored at the last observed frame.}
\label{app:conditioned-dynamics-example}
\end{figure*}

\begin{figure*}[t]
\centering
\includegraphics[width=\textwidth]{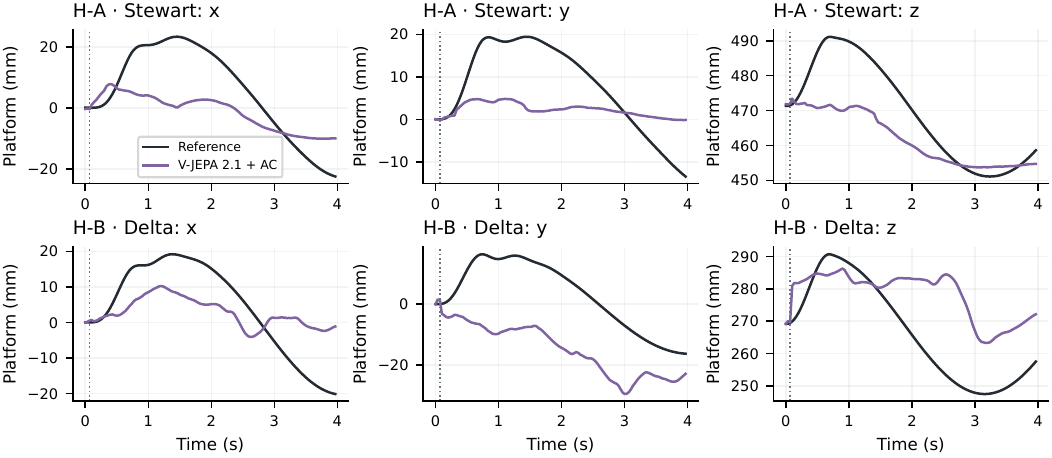}
\caption{Spatial platform trajectories for the first registered Stewart and Delta test cases, seed 42. Black curves are reference positions; purple curves are V-JEPA 2.1 + AC with the frozen physical readout. The dotted line ends the three-frame observed prefix. All three coordinates and the full forecast horizon are shown.}
\label{app:spatial-traces}
\end{figure*}

\section{Spatial adaptation and independent SI evaluation}
\label{app:spatial}
\subsection{Data, representation, and training}
The spatial studies share 50 Stewart (H-A) and 50 Delta (H-B) test assemblies, each with three operating conditions, and use seeds 42, 43, and 44. The resulting 300-trajectory cohort is common to V-JEPA + AC and LingBot; each model retains its native training objective and budget. Assembly-level training/validation/test partitions are disjoint. Stewart has 14 training and four validation identities, while Delta has four of each. All five training views are used, giving 210 and 60 videos; validation and testing use the front view, giving 12 validation and 150 test trajectories per family. The assembly is the independent sampling unit. Each video contains 120 frames at 30 Hz, rendered at $640\times640$ and resized to $384\times384$ for V-JEPA's encoder.

The spatial model combines the released V-JEPA 2.1-B/16 encoder~\cite{murlabadia2026vjepa2_1} with the AC predictor architecture~\cite{assran2025vjepa2}. The frozen encoder has 86.8 million parameters. Each observed frame forms a duplicated-frame tubelet, yielding 576 spatial tokens of width 768. A separate 24-layer predictor, width 1024 with 16 heads, is initialized and trained for each family. Its inputs are visual features, standardized actuator position/velocity commands, and actuator state feedback. Training uses true training-state feedback; forecasting uses the model's own decoded state.

A physical readout pools encoder features to a $4\times4$ grid, concatenates 13 public camera-calibration values, and applies hidden layers of widths 512 and 256. Its outputs are the complete tree generalized position and velocity: 25/24 coordinates for Stewart and 22/21 for Delta. Training uses standardized state MSE for 60 epochs, AdamW with learning rate $3\times10^{-4}$ and weight decay $10^{-4}$, and batch size 256. Only the free-root quaternion is normalized. This seed-specific head supplies recurrent state feedback during forecasting; the independently calibrated heads below provide the common measurement interface. Their comparison holds the entire predicted feature trajectory fixed.

The AC predictor uses eight-frame clips, teacher-forced feature L1 loss plus a two-step autoregressive L1 loss, and AdamW with weight decay 0.04 and maximum learning rate $4.25\times10^{-4}$. Each family and seed receives 3,000 sequence exposures with batch size eight and final-checkpoint forecasting: Stewart uses 386 updates over 15 epochs across 210 videos, with 60 exposures in the last epoch; Delta uses 400 updates over 50 epochs across 60 videos. The learning rate follows a cosine schedule fixed at initialization for the complete 15/50-epoch run, with warmup for the first 5\% of epochs (at least one epoch). Parameters use float32 and forward arithmetic uses bfloat16 autocast.

At test time, the model observes three RGB frames and predicts the remaining 117 at 30 Hz with a seven-frame autoregressive context, using public camera calibration and planned actuator commands. The audit measures this combined encoder, predictor, and physical-readout pathway. Because its outputs are discrete latent features, physical states are the common evaluation interface; RGB reconstruction and continuous-time ODE solvers belong to the other model pathways. Figure~\ref{app:spatial-traces} shows the first registered test trajectory of each family.

\begin{figure*}[t]
\centering
\includegraphics[width=\textwidth]{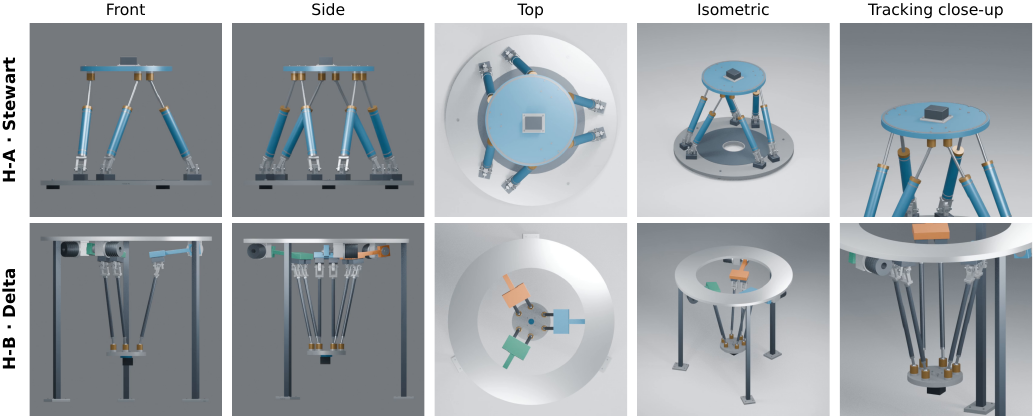}
\caption{Five training camera views of Stewart (top) and Delta (bottom). Each row shows frame zero from the first registered training assembly in its low-speed condition. All views show the same physical state; only the camera changes. The full recorded frames are shown without cropping.}
\label{app:spatial-overview}
\end{figure*}

\subsection{Reaction-free forces in multiple dimensions}
Spatial closure is represented by attachment-point gaps $\phi(q)\in\R^{3m}$ and velocity Jacobian $J$. The free-root quaternion uses its tangent angular velocity, so the velocity dimension is one smaller than the stored position dimension. With generalized acceleration $a$, the rigid-body equation is
\begin{equation}
Ma+c+G=Bu_A+Q_{\rm ext}+Q_f+J\T\lambda.
\end{equation}
An orthonormal basis $N$ of $\ker J$ eliminates $\lambda$. At a regular configuration where $N\T B$ is square and nonsingular, the actuator-equivalent residual follows from
\begin{equation}
N\T B\,r_u=N\T(M\hat a+c+G-Q_{\rm ext}-Q_f)-N\T B\,u_A.
\end{equation}
The basis choice cancels in the solution: replacing $N$ by $NR$ for invertible $R$ multiplies both sides by $R\T$. All rigid-body terms are constructed independently from the reference geometry and body properties. The position servo, including its supplied feedforward term and saturation, is reevaluated at the predicted actuator position and velocity. Stewart's six prismatic drives give force residuals in N; Delta's three rotary drives give torque residuals in N\,m. Physical energy uses the same body model, and power is $(Bu_A+Q_{\rm ext}+Q_f)\T\hat v$.

\paragraph{Metrics and calibration.}
Closure, $J\hat v$, and $J\hat a+\dot J\hat v$ use the maximum Euclidean attachment-gap norm over points and evaluated times. Their tolerances are 1 mm, 0.01 m/s, and 0.1 m/s$^2$. Point error is Euclidean RMSE over all future samples and the specified markers: platform center plus rod origins for Stewart; platform center, elbows, and distal rod endpoints for Delta. Its tolerance is 10 mm. Effort RMSE averages actuator channels and evaluated times, with thresholds 7.5 N and 0.75 N\,m. The energy threshold is $0.01u_{\max}\sum_j\operatorname{range}(q_{{\rm cmd},j})$, using each registered command trajectory.

Generalized velocities use centered configuration differences; a seven-frame, degree-three Savitzky--Golay filter smooths velocities and differentiates them for acceleration. Point prediction uses frames 3--119 (zero-based); differentiated mechanics uses frames 3--116, and work is anchored at frame 3. All 150 reference trajectories per family meet the six criteria with either exact or 30 Hz sampled derivatives. For sampled derivatives, maximum effort errors are 0.0765 N for Stewart and 0.0220 N\,m for Delta, with maximum energy defects of 0.0530 and 0.0681 J. Exact recorded derivatives give maximal effort residuals of $6.20\times10^{-14}$ N and $1.11\times10^{-14}$ N\,m, respectively.

\paragraph{Reconstruction and coverage.}
The post-hoc intervention retains Stewart's predicted platform pose or Delta's predicted platform position and reconstructs the legs on the declared branch after the complete rollout. CAD/IK reconstruction succeeds on 450/450 Stewart and 450/450 Delta forecasts; differentiated-constraint checks leave 450 and 432 mechanically measurable cases. Mean coordinate corrections are 2.32/2.64 mm for Stewart/Delta, and reconstructed point errors are 31.75/25.68 mm (Table~\ref{app:spatial-ac-seeds}). SI residuals average mechanically measurable cases within each seed and then weight seeds equally.

\paragraph{Paired prediction and interpretation.}
The mean native-minus-persistence point error is $+2.57$ mm for Stewart and $+4.49$ mm for Delta. The corresponding 95\% intervals are $[-0.18,6.64]$ and $[2.21,7.62]$ mm. We use 10,000 hierarchical bootstrap draws, resampling the three training seeds and 50 test identities while retaining all three conditions of each identity together. The readout contrasts below hold the predicted feature trajectories fixed, isolating how coordinate recovery changes these forecasts' physical measurements.

The public camera framing uses the source trajectory bounds, and controller feedforward is supplied as part of the calibrated command rule. Spatial results use their own actuator scales and sampling definitions alongside the planar study's common diagnostic categories.

\begin{table}[t]
\caption{V-JEPA + AC on 50 test assemblies per family, three conditions and three seeds. Scores give the mean $\pm$ seed SD. CAD reconstruction is a separate intervention.}
\label{app:spatial-ac-scores}
\centering\small
\begin{tabular}{lrrrr}
\toprule
Family & Native $S_{\rm comp}$ & CAD $S_{\rm comp}$ & Mech. & Joint\\
\midrule
Stewart & $6.20\!\pm\!0.69$ & $55.11\!\pm\!1.71$ & 450/450 & 0/450 \\
Delta & $10.40\!\pm\!5.34$ & $50.05\!\pm\!2.71$ & 432/450 & 0/450 \\
\bottomrule
\end{tabular}
\end{table}

\begin{table*}[t]
\caption{V-JEPA + AC by seed on 150 trajectories per family. Point errors and correction costs are in mm; effort is in N for Stewart and N\,m for Delta, and energy is in J. CAD residuals average mechanically measurable cases.}
\label{app:spatial-ac-seeds}
\centering\small
\setlength{\tabcolsep}{4pt}
\begin{tabular}{llrrrrrrrr}
\toprule
Family & Seed & Native $S_{\rm comp}$ & CAD $S_{\rm comp}$ & Mech. & Correction & Raw point & CAD point & Effort & Energy\\
\midrule
Stewart & 42 & 6.51 & 54.26 & 150/150 & 2.36 & 29.00 & 29.01 & 39.652 & 2.140 \\
Stewart & 43 & 6.68 & 57.07 & 150/150 & 1.95 & 30.47 & 30.15 & 42.015 & 0.971 \\
Stewart & 44 & 5.41 & 53.99 & 150/150 & 2.65 & 36.02 & 36.10 & 44.235 & 1.400 \\
Delta & 42 & 10.98 & 50.52 & 147/150 & 2.13 & 23.47 & 23.44 & 12.570 & 2.688 \\
Delta & 43 & 4.78 & 47.14 & 135/150 & 2.38 & 29.22 & 29.00 & 14.893 & 1.342 \\
Delta & 44 & 15.42 & 52.50 & 150/150 & 3.40 & 24.93 & 24.58 & 9.812 & 3.356 \\
\bottomrule
\end{tabular}
\end{table*}

\subsection{Frozen-feature readout intervention}
\label{app:ac-readout}
To separate coordinate recovery from feature forecasting, we decode all 900 saved AC rollouts with both the internal feedback head and the fixed calibrated measurement head. The internal head provides all recurrent state feedback throughout the rollout. All readout contrasts use the same saved feature trajectories from FP32 checkpoints with BF16 forecast arithmetic, and alternative heads are applied after the complete rollout. Correct-video features pass through the same heads, CAD reconstruction, and 30 Hz differentiator as predicted features, so each mechanical comparison pairs the same trajectory and measurement path.

The CAD-coordinate head estimates offsets from the public nominal configuration, using the training protocol in Section~\ref{app:lingbot}. For Delta, a reset-relative linear head instead maps feature changes from the first observed frame to coordinate changes from the known reset pose. It is fitted on 12 front-view training clips at 384 pixels. Six disjoint calibration clips select ridge regularization from $\{10^{-4},10^{-3},10^{-2},10^{-1},1\}$ by coordinate, velocity, and acceleration error; six further clips provide independent verification. This readout retains framewise coordinates without smoothing or geometric projection, and its objective uses coordinate supervision. Inverse kinematics remains a separate intervention applied to every head. Table~\ref{app:ac-readout-table} uses 450 forecasts and 150 distinct reference videos per family; the seed-specific internal heads each replay these reference videos.

\begin{table*}[t]
\caption{Three independently trained AC predictors with internal and calibrated physical readouts. Reference and forecast features share each measurement head. Point errors are in mm; effort is in N for Stewart and N\,m for Delta, and energy is in J. Raw/CAD scores distinguish unconstrained decoded coordinates from their subsequent CAD/IK reconstruction. SI residuals average the mechanically measurable cases.}
\label{app:ac-readout-table}
\centering\small
\setlength{\tabcolsep}{4pt}
\begin{tabular}{lllrrrrrrr}
\toprule
Family & Readout & Input & \multicolumn{2}{c}{Point RMSE} & \multicolumn{2}{c}{$S_{\rm comp}$} & \multicolumn{3}{c}{After CAD reconstruction}\\
\cmidrule(lr){4-5}\cmidrule(lr){6-7}\cmidrule(l){8-10}
& & & Raw & CAD & Raw & CAD & Mech. & Effort & Energy\\
\midrule
Stewart & Internal & Reference & 3.59 & 4.52 & 15.63 & 71.32 & 450/450 & 8.882 & 2.186 \\
Stewart & Calibrated & Reference & 2.07 & 1.88 & 28.88 & 86.46 & 150/150 & 3.108 & 0.627 \\
Stewart & Internal & Forecast & 33.11 & 44.59 & 1.66 & 43.51 & 343/450 & 72.865 & 11.354 \\
Stewart & Calibrated & Forecast & 31.83 & 31.75 & 6.20 & 55.11 & 450/450 & 41.967 & 1.504 \\
Delta & Internal & Reference & 10.29 & 19.23 & 10.10 & 36.97 & 298/450 & 9.335 & 4.948 \\
Delta & Calibrated & Reference & 3.59 & 3.32 & 27.85 & 67.11 & 150/150 & 2.072 & 0.990 \\
Delta & Internal & Forecast & 27.10 & 38.29 & 2.78 & 38.09 & 287/450 & 14.648 & 6.429 \\
Delta & Calibrated & Forecast & 25.87 & 25.68 & 10.40 & 50.05 & 432/450 & 12.425 & 2.462 \\
\bottomrule
\end{tabular}
\end{table*}

Correct-video closure residuals average 2.41/3.12 mm for Stewart/Delta before reconstruction, locating part of the native admissibility gap in coordinate recovery. Matched replay then quantifies the measurement floor for force and energy: Table~\ref{app:ac-reference} pairs reference and forecast residuals on the same mechanically measurable reconstructed trajectories. The fixed-feature readout contrast isolates the contribution of coordinate recovery.

\begin{table}[t]
\caption{Matched AC reference and forecast mechanics after calibrated readout and CAD/IK reconstruction. Each mean first averages the measurable cases within a seed, then averages seeds. Effort is in N for Stewart and N\,m for Delta; energy is in J.}
\label{app:ac-reference}
\centering\footnotesize
\setlength{\tabcolsep}{3pt}
\begin{tabular}{llrrrr}
\toprule
& & \multicolumn{2}{c}{Effort} & \multicolumn{2}{c}{Energy}\\
Family & Pairs & Ref. & Pred. & Ref. & Pred.\\
\midrule
Stewart & 450 & 3.108 & 41.967 & 0.627 & 1.504 \\
Delta & 432 & 2.083 & 12.425 & 0.985 & 2.462 \\
\bottomrule
\end{tabular}
\end{table}

\subsection{LingBot-VA generation and fixed-budget adaptation}
\label{app:lingbot}
\paragraph{Cohort and model.}
The generative study evaluates 50 Stewart and 50 Delta test assemblies. Each identity has low-speed, rated, and increased-load trajectories, giving 150 tests per family and 300 in total. Assembly identities are disjoint from the training and validation partitions; the three operating conditions remain grouped by identity. The same 210 Stewart and 60 Delta training videos cover five camera views, while testing uses the front view.

We use the released LingBot-VA base model~\cite{li2026causal}, whose shared Transformer contains 5,088,872,670 parameters. Zero-shot generation uses the public base weights. Adaptation updates all Transformer parameters with the native joint video/action flow-matching objective, while freezing the VAE and text encoder. Each 120-frame training video is resized to $256\times256$; repeating its last frame gives 121 frames for VAE encoding into 31 latent frames.

\paragraph{Training and replication.}
Each of seeds 42, 43, and 44 starts independently from the pretrained weights and completes 50 updates. AdamW uses learning rate $10^{-5}$, $\beta=(0.9,0.95)$, weight decay 0.1, and $\epsilon=10^{-8}$, with ten warmup updates. Microbatch size one and eight-step gradient accumulation give effective batch size eight and 400 sequence exposures per seed. Parameters and optimizer states use FP32, with BF16 forward arithmetic. We evaluate the prescribed 0- and 50-update checkpoints. Table~\ref{app:lingbot-seeds} reports all three repetitions; score dispersion is the sample standard deviation across seeds.

\paragraph{Native causal generation and command interface.}
Each test uses the last image of the common three-frame prefix as LingBot's single-image initialization. The remaining 117 frames are generated at 30 Hz and $256\times256$ using eight chunks of four latent frames, five video denoising steps, and classifier-free guidance scale five. Public actuator position commands are resampled to 60 Hz and normalized by the training set's first and 99th percentiles for each family. Stewart occupies action channels 14--19 and Delta channels 14--16, with zero-based indexing; unused channels are zero-filled and masked in the training loss. The native causal order generates each video chunk before inserting its known planned commands into the action cache for subsequent chunks. Generation seeds are paired by training-seed/test-case identity across checkpoints. The three zero-shot repetitions share the base weights and vary generation randomness; the adapted repetitions additionally vary training randomness. All 1,800 videos are finite, comprising 210,600 generated future frames.

\paragraph{Calibrated RGB readout.}
Both reference and generated images enter the same frozen V-JEPA 2.1-B/16 encoder at 384 pixels. Stewart uses a CAD-coordinate head trained for 240 epochs on front-view training videos at 256 and 384 pixels, with AdamW, learning rate $3\times10^{-4}$ decaying to $3\times10^{-6}$, and weight decay $10^{-4}$. Its six-clip verification point RMSE is 2.65 mm.

Delta uses the public CAD reset pose and the observed first frame to anchor absolute position. A linear ridge head maps differences from the first-frame features to unconstrained coordinate offsets; training uses the 12 original front-view training videos. Six calibration clips select the ridge coefficient from $\{10^{-4},10^{-3},10^{-2},10^{-1},1\}$ by position and derivative errors, choosing $10^{-2}$ after kernel-trace normalization. Six disjoint verification clips assess the frozen configuration. No force, energy, or closure objective enters this fit.

For the orthographic camera, platform surface correspondences provide motion in the image plane, while feature regression supplies depth along the camera axis. Nominal CAD visibility defines candidate surface pixels; pyramidal Lucas--Kanade tracking uses a nine-pixel window, three pyramid levels, and a 0.35-pixel forward--backward check. The median displacement of at least four valid correspondences determines image-plane translation. Coordinates remain framewise; the registered derivative estimator is applied during auditing. Native coordinates are scored before the separate CAD/IK reconstruction.

\paragraph{Image prediction and persistence.}
Whole-image RGB MSE averages all pixels and channels in $[0,1]$ over the 117 future frames, then weights test cases equally. The motion mask contains pixels whose reference future RGB differs from the last observed image by more than $2/255$ in any channel. Last-frame image persistence gives MSE $29.92/13.73\times10^{-4}$ and motion-mask MSE $2.512/1.447\times10^{-2}$ for Stewart/Delta. Both checkpoints exceed these baselines. Adaptation lowers whole-image MSE by 12.6\%/2.1\%, while motion-mask MSE increases by 4.0\%/7.2\%. The improvement in global image similarity therefore does not extend to the moving foreground.

Point persistence freezes the last physical readout of each sequence. Figure~6c in the main paper plots every finite forecast's point error minus its own persistence error. The means remain close to zero for Stewart and positive for Delta; reconstructed point RMSE at 50 updates is 29.92/23.24 mm. These paired comparisons retain identical future time supports and coordinate definitions.

\begin{table*}[t]
\caption{LingBot results by seed on 150 test trajectories per family. RGB MSE uses $10^{-4}$ units and motion-mask MSE $10^{-2}$. Force and energy means use the mechanically measurable CAD subset. Effort is in N for Stewart and N\,m for Delta; energy is in J.}
\label{app:lingbot-seeds}
\centering\small
\setlength{\tabcolsep}{4pt}
\begin{tabular}{llrrrrrrrr}
\toprule
Family & Updates & Seed & RGB & Motion & Native $S_{\rm comp}$ & CAD $S_{\rm comp}$ & Mech. & Effort & Energy\\
\midrule
Stewart & 0 & 42 & 38.84 & 2.55 & 5.64 & 54.38 & 150/150 & 39.548 & 1.153 \\
Stewart & 0 & 43 & 48.55 & 2.67 & 5.73 & 54.65 & 150/150 & 39.595 & 1.114 \\
Stewart & 0 & 44 & 37.62 & 2.54 & 5.65 & 54.75 & 150/150 & 39.401 & 1.074 \\
Stewart & 50 & 42 & 35.23 & 2.63 & 6.15 & 54.88 & 150/150 & 40.499 & 1.354 \\
Stewart & 50 & 43 & 37.19 & 2.72 & 5.85 & 53.88 & 150/150 & 40.278 & 1.931 \\
Stewart & 50 & 44 & 36.85 & 2.72 & 6.02 & 54.17 & 150/150 & 40.763 & 1.635 \\
Delta & 0 & 42 & 28.20 & 1.59 & 2.71 & 42.03 & 116/150 & 12.232 & 0.622 \\
Delta & 0 & 43 & 27.42 & 1.56 & 2.86 & 44.19 & 122/150 & 12.279 & 0.640 \\
Delta & 0 & 44 & 32.67 & 1.60 & 2.84 & 43.89 & 121/150 & 12.374 & 0.577 \\
Delta & 50 & 42 & 26.87 & 1.65 & 3.40 & 53.10 & 143/150 & 12.649 & 0.578 \\
Delta & 50 & 43 & 30.75 & 1.72 & 3.41 & 50.77 & 139/150 & 11.994 & 1.999 \\
Delta & 50 & 44 & 28.86 & 1.72 & 3.36 & 51.90 & 142/150 & 12.315 & 1.040 \\
\bottomrule
\end{tabular}
\end{table*}

\paragraph{Reconstruction and coverage.}
The frozen measurement path, engineering targets, and CAD/IK intervention are common to both checkpoints. Every available native readout violates closure, so native mechanics coverage is zero. Stewart supplies 450/450 finite native readouts and mechanically measurable reconstructions at each checkpoint. Delta's finite native readouts increase from 374 to 442 out of 450. Its reachable reconstructions increase from 372 to 428, and differentiated-constraint checks leave 359 and 424 mechanically measurable cases.

Table~\ref{app:lingbot-components} separates correction cost from the six reconstructed contributions. Stewart's reconstructed $S_{\rm comp}$ stays near 54; Delta rises from $43.37\pm1.17$ to $51.92\pm1.17$. The latter gain is concentrated in coverage and kinematic consistency: its effort contribution remains below 0.4 out of 100. Strict joint qualification is zero for every seed, family, checkpoint, and measurement track. Reference replay below measures the image-to-mechanics instrument under the same conditions.

\begin{table*}[t]
\caption{LingBot reconstruction and score components. Counts pool three seeds (450 registered cases per row). Correction and point errors are in mm and average available cases within seed, then across seeds. $G,V,A,F,E,P$ are reconstructed closure, velocity, acceleration, effort, energy, and point scores on a 0--100 scale; their mean is $S_{\rm comp}$.}
\label{app:lingbot-components}
\centering\small
\setlength{\tabcolsep}{3.5pt}
\begin{tabular}{llrrrrrrrrrrr}
\toprule
Family & Updates & Readout & Reach. & Correction & Point & CAD point & $G$ & $V$ & $A$ & $F$ & $E$ & $P$\\
\midrule
Stewart & 0 & 450/450 & 450/450 & 2.08 & 29.68 & 29.63 & 100.00 & 99.59 & 96.68 & 3.49 & 17.50 & 10.32 \\
Stewart & 50 & 450/450 & 450/450 & 2.11 & 30.01 & 29.92 & 100.00 & 99.76 & 98.32 & 3.37 & 14.14 & 10.26 \\
Delta & 0 & 374/450 & 372/450 & 7.64 & 22.80 & 22.01 & 82.67 & 80.34 & 71.43 & 0.30 & 11.15 & 14.34 \\
Delta & 50 & 442/450 & 428/450 & 9.06 & 23.23 & 23.24 & 95.11 & 94.23 & 90.38 & 0.37 & 15.88 & 15.58 \\
\bottomrule
\end{tabular}
\end{table*}

\section{Continuous scoring and measurement calibration}
\label{app:measurement}
\subsection{One continuous score at fixed engineering targets}
Each criterion reduces a trajectory to its registered nonnegative SI error, which the continuous map $s(\rho)=100/(1+\rho^2)$ scores using $\rho=r/\tau$. For multiple channels, $\rho$ is the largest channel-wise error-to-tolerance ratio. The primary $S_{\rm comp}$ averages all six contributions over registered trajectories with equal condition, family, and seed weights. Missing measurements contribute zero to this fixed denominator and case count, while their raw physical errors are reported as N/A.

The map assigns 100, 50, and 20 points at errors of zero, one, and two tolerances, respectively, capturing improvements such as KeyCLD's RK4 gains even when strict joint qualification remains zero. Three perfectly satisfied criteria contribute 50 points to the composite, while raw residuals, physical coverage, and the joint rate show which requirements still fail. Native predictions and reconstructed trajectories use the same scoring rule, with their results reported separately to expose the effect of correction.

Reconstruction introduces public geometry into saved predictions. In the overview figure, KeyCLD at 256 pixels receives 50.00 points from reconstructed closure and its two differentiated constraints, and 3.10 from effort, energy, and tracking combined. Adapted LingBot receives 48.15 and 4.97 points from these respective groups. Their similar totals, 53.10 and 53.12, thus reflect different component profiles on different tasks. The planar and spatial targets normalize each task's engineering requirements; model comparisons within a task retain the same mechanisms, commands, and targets.

Engineering targets describe the task, mechanism, and actuator. They remain fixed across models and input resolutions within a task; spatial effort and work targets use the registered actuator and command scales. Measurement calibration specifies the camera, reader, derivative estimator, and operating range. Instruments are fitted on training identities, processing is selected on development references, and correct videos pass through the same frozen path as generated videos. Paired reference and forecast residuals retain the measured SI values and their common engineering targets.

For a calibrated horizontal field of view $L$, pixel spacing is $L/N$ when $L$ spans pixel edges, or $L/(N-1)$ when it spans first and last pixel centers. The planar corpus uses the latter convention with $L=1.30$ m. Pixel spacing converts image positions into meters, whereas localization accuracy depends on the reader and can be subpixel. Coordinate bias, covariance, temporal dependence, and the nonlinear measurement path govern propagation into velocity, acceleration, effort, and work. Hard geometric constraints independently determine rod-length preservation.

\paragraph{Sampling variation and reference calibration.}
Seed/assembly bootstrap intervals describe variation in model results under a fixed measurement path, while calibration quantiles characterize that path's reference errors. Whole-trajectory summaries preserve coordinate bias, centered spread, and temporal dependence. For the CAD-coordinate head, empirical 95th percentiles from six calibration clips per family are 2.39/3.79 mm on Stewart/Delta and cover 2/6 and 3/6 verification clips. This verification coverage measures the transfer of empirical error levels to new identities, complementing the bootstrap intervals on model performance.

\subsection{64-pixel reference replay and independent readouts}
The planar calibration partition contains 18 identities and 54 clips. We divide it by identity into nine calibration and nine verification mechanisms, retaining all three conditions per identity. Spatial readout calibration uses the eight original validation identities: four for calibration and four for verification, giving 12 clips in each group. Each spatial family therefore contributes six verification clips. We replay all 90 planar or 120 spatial frames, retain the registered future interval, and measure differentiation on a common interior support. Trajectory-level summaries retain temporal dependence. Development comparisons use these held-out references; frozen reference replay evaluates the 100 spatial test identities at each model's input resolution.

KeyCLD's permutation and CON's affine map are parts of their declared model interfaces. To ablate their influence, an independent planar image reader predicts three unconstrained heatmaps from RGB alone. Its U-Net uses widths 16/32/64/96, GroupNorm, and soft-argmax coordinates. Training uses the 813 original training sequences, 32 epochs, batch 128, and AdamW with learning rate $0.002$ decaying to $0.0002$. It receives no geometry or physical constraints. Eight further epochs add a balanced mixture of clean and degraded training frames: Gaussian blur with $\sigma\in[0.2,1.5]$, contrast in $[0.5,1.1]$, and noise of $1/255$ standard deviation. Both checkpoints are evaluated on the same verification frames.

\begin{table}[t]
\caption{Planar reader ablation on 27 verification sequences. Entries are mean Euclidean point RMSE (mm). Augmentation targets image degradation without imposing closure.}
\label{app:reader-quality}
\centering\small
\begin{tabular}{lrr}
\toprule
Reference image quality & Clean-trained & Augmented\\
\midrule
Clean & 1.81 & 1.90 \\
Blur $\sigma=0.5$ & 2.92 & 2.04 \\
Blur $\sigma=1$ & 13.22 & 2.64 \\
Blur + $0.6\times$ contrast & 64.80 & 2.73 \\
\bottomrule
\end{tabular}
\end{table}

Augmentation reduces degradation-induced errors while preserving about 2 mm clean-image precision (Table~\ref{app:reader-quality}). Applying this frozen reader to saved forecasts and paired correct videos reveals large forecast coordinate errors alongside 2.13 mm reference localization error through the same measurement path (Table~\ref{app:reader-paired}).

\begin{table}[t]
\caption{Matched public-reader ablation, 108 planar test clips per seed. Point: Euclidean RMSE from the common frozen RGB reader; torque and energy: the projected phase diagnostic below. Means use finite cases. Reference replay measures the instrument's own residuals.}
\label{app:reader-paired}
\centering\small
\setlength{\tabcolsep}{3pt}
\begin{tabular}{lrrrr}
\toprule
RGB source & Finite & \shortstack{Point\\(mm)} & \shortstack{Torque\\(N\,m)} & \shortstack{Energy\\(J)}\\
\midrule
Reference RGB & 108/108 & 2.13 & 0.73 & 0.57 \\
KeyCLD, 42 & 108/108 & 185.68 & 7.00 & 29.87 \\
KeyCLD, 43 & 108/108 & 183.16 & 6.69 & 8.06 \\
KeyCLD, 44 & 95/108 & 191.94 & 6.86 & 23.96 \\
CON, 42 & 108/108 & 167.70 & 6.49 & 7.46 \\
CON, 43 & 108/108 & 165.23 & 6.91 & 19.38 \\
CON, 44 & 108/108 & 168.26 & 6.67 & 11.30 \\
\bottomrule
\end{tabular}
\end{table}

\subsection{Numerical differentiation and sensitivity controls}
Exact sampled trajectories isolate differentiation from visual localization. On 54 independent 30 Hz references, a seven-frame cubic Savitzky--Golay estimator passes all six criteria on 40/54 trajectories. A quintic interpolating spline passes 54/54 on the same trajectories. Worst-case velocity and acceleration constraint errors fall from $0.0219$ m/s and $2.119$ m/s$^2$ to $0.000281$ m/s and $0.0249$ m/s$^2$. Its worst-case torque and energy errors are $0.00298$ N\,m and $0.01291$ J. A 5 mm radial displacement, $0.8\times$ time scaling, and a $0.5$ N\,m force offset remain detectable in all 54 controls.

Differentiating image-derived coordinates also amplifies localization noise. For the geometric intervention only, we fit crank phase locally and reconstruct derivatives analytically on the manifold. We select window length and polynomial degree by development position, velocity, and acceleration error, subject to all 54 exact-state development trajectories still passing. This selects a seven-frame cubic phase estimator; a 21-frame cubic fit lowers image-derived residuals but fails the exact-state preservation requirement. Native point and closure errors are computed before this intervention.

On the 27 clean verification clips, the phase estimator reduces mean reference-image torque error from 1.94 to 0.74 N\,m and energy error from 2.25 to 0.51 J. It preserves 54/54 exact-state passes and detects the time-scaling, 1 Hz phase, and force-offset controls. Its acceleration gain falls from 0.965 at 1 Hz to 0.720 at 3 Hz, motivating joint reporting with the unfiltered coordinate diagnostic. The remaining correct-video force and energy errors exceed engineering targets, so Table~\ref{app:reader-paired} characterizes the combined forecasting and RGB measurement path. Native-ODE derivatives and exact-state controls provide the complementary tests that isolate constitutive response from visual localization.

\begin{figure*}[t]
\centering
\includegraphics[width=.96\textwidth]{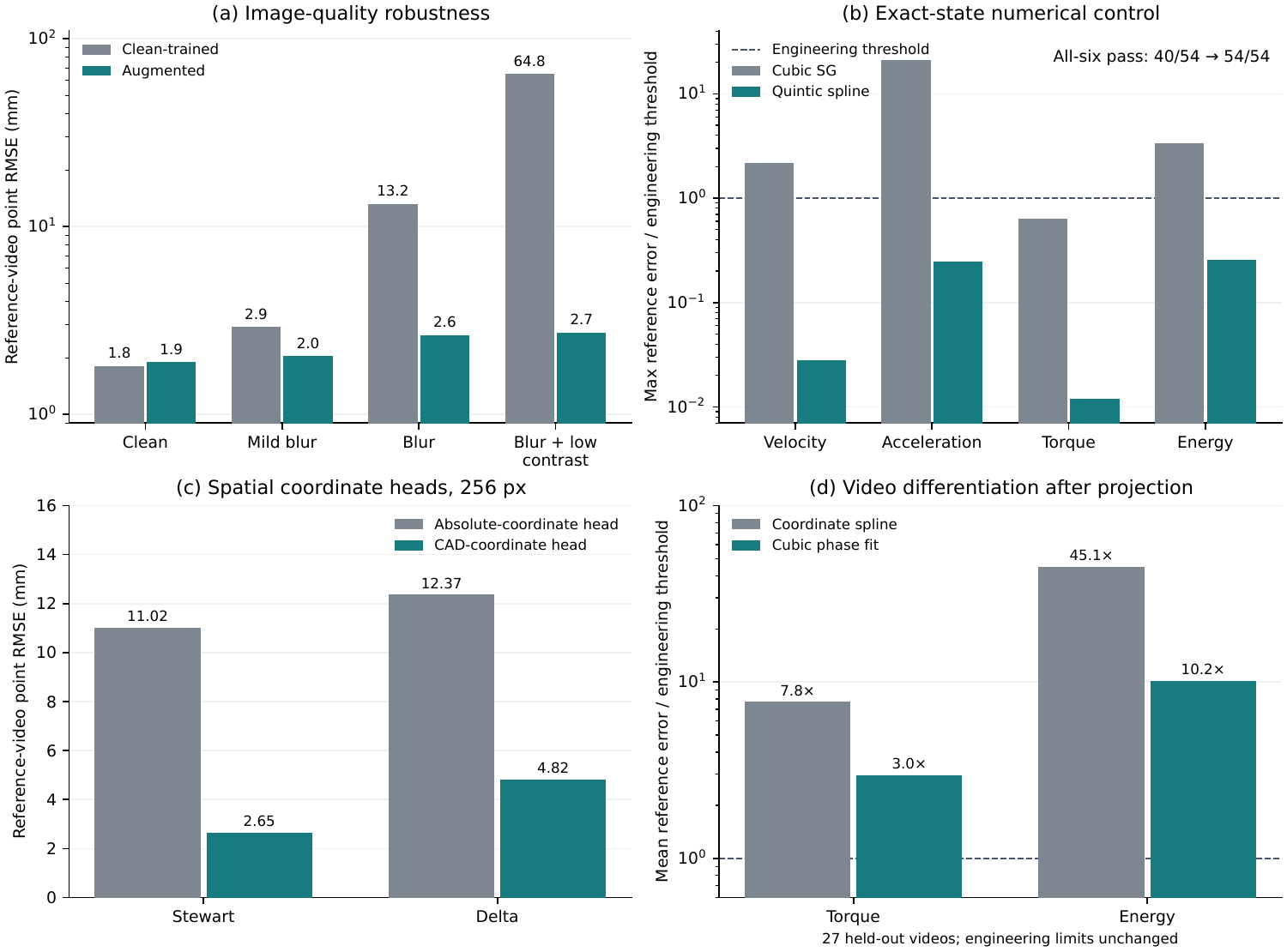}
\caption{Measurement controls. (a) Planar reader training with and without degradation augmentation. (b) Exact-coordinate differentiation errors relative to engineering targets. (c) Absolute-coordinate and CAD-coordinate heads at 256-pixel input resolution. (d) Coordinate-spline and cubic-phase differentiation of correct videos. Identical reference trajectories separate measurement accuracy from forecasting performance; configuration details and sensitivity controls are given in the text.}
\label{app:calibration-figure}
\end{figure*}

\subsection{Spatial coordinate-head comparisons}
We compare two unconstrained coordinate readers on the same 256-pixel verification videos (Table~\ref{app:observer-ablation}). The absolute-coordinate head uses the state-regression configuration in Section~\ref{app:spatial}: 60 training epochs and all five training views. The CAD-coordinate head predicts free offsets from a public CAD origin, using front-view training at two input resolutions for 240 epochs (Section~\ref{app:lingbot}). Both share frozen V-JEPA features and use no closure loss or inverse-kinematics projection. This compares complete measurement configurations, including their training protocols. A third condition aligns the CAD-coordinate output with a known observed reset pose. The Stewart audit uses the CAD-coordinate head. Delta combines reset-relative feature regression with image-plane tracking, using the known reset pose; Table~\ref{app:delta-components} separates these components.

\begin{table}[t]
\caption{Coordinate-head verification, six clips per family. Point RMSE is in mm; coverage counts mechanically measurable reconstructions. The alignment row additionally uses a known initial pose.}
\label{app:observer-ablation}
\centering\small
\setlength{\tabcolsep}{3pt}
\begin{tabular}{lrrrr}
\toprule
Configuration & Stewart & Delta & S cov. & D cov.\\
\midrule
Absolute-coordinate head & 11.02 & 12.37 & 6/6 & 3/6 \\
CAD-coordinate head & 2.65 & 4.82 & 6/6 & 6/6 \\
+ known-pose alignment & 2.96 & 3.44 & 6/6 & 6/6 \\
\bottomrule
\end{tabular}
\end{table}

\paragraph{Delta measurement components.}
The independent verification set isolates the contributions of reset-relative regression and image-plane tracking (Table~\ref{app:delta-components}). Tracking reduces torque residuals from 2.14 to 0.81 N\,m and work--energy defects from 0.83 to 0.18 J. Exact sampled-state controls give 0.014 N\,m and 0.023 J, separating numerical differentiation from image localization. A held-reset negative control has zero work defect but 12.13 N\,m torque error, illustrating why the effort and energy criteria complement each other.

\begin{table}[t]
\caption{Delta readout components on six independent verification clips. All rows use the same CAD reconstruction and derivative estimator. Point is full-marker Euclidean RMSE in mm; effort is in N\,m and energy in J. The held reset is a coordinate persistence control.}
\label{app:delta-components}
\centering\small
\setlength{\tabcolsep}{3pt}
\begin{tabular}{lrrr}
\toprule
Input / configuration & Point & Effort & Energy\\
\midrule
Exact sampled states & 0.00 & 0.01 & 0.02 \\
Relative feature head & 3.54 & 2.14 & 0.83 \\
+ image-plane tracking & 2.44 & 0.81 & 0.18 \\
Held CAD reset & 21.22 & 12.13 & 0.00 \\
\bottomrule
\end{tabular}
\end{table}

\paragraph{Matched reference replay.}
All 300 correct videos pass through the same frozen family-specific readout, CAD reconstruction, and derivative estimator as the generated videos. Reference reconstructions are mechanically measurable on all 150 trajectories per family. Their point RMSE is 1.83 mm for Stewart and 2.25 mm for Delta, compared with 29.92/23.24 mm for adapted generated videos. Exact-state and sampled-state controls quantify the numerical contribution independently of RGB localization (Table~\ref{app:spatial-controls}).

Table~\ref{app:spatial-reference} pairs reference and generated mechanical residuals on identical cases within each seed. At 50 updates, generated effort error exceeds reference error on all 450 Stewart and 424 Delta pairs. The means are 3.009 versus 40.513 N and 0.733 versus 12.319 N\,m. Delta's sampled-state differentiation control contributes 0.0145 N\,m, so the large generated torque discrepancy is well separated from both numerical and correct-video residuals.

Matched reference/generated work--energy defects are 0.605/1.640 J for Stewart and 0.203/1.206 J for Delta; generated defects are larger on 390/450 and 330/424 pairs. Delta's 0.203 J paired reference residual includes RGB recovery, CAD reconstruction, and differentiation; sampled states give 0.0253 J (Table~\ref{app:spatial-controls}). Reference RGB energy passes are 48/150 Stewart and 63/150 Delta trajectories, with exact and sampled states passing throughout. The trajectory-specific spatial targets span 0.343--0.504 J for Stewart and 0.106--0.172 J for Delta. Zero- and 50-update Delta means use 359 and 424 measurable forecasts, respectively. Section~\ref{app:ac-readout} gives the frozen-feature readout intervention for V-JEPA + AC.

\begin{table}[t]
\caption{Matched LingBot mechanics after CAD reconstruction. Reference and generated videos use identical cases and processing. Values average per-seed available-case means; $n$ pools the three seeds out of 450 registered cases. Effort is in N (Stewart) or N\,m (Delta); energy is in J.}
\label{app:spatial-reference}
\centering\small
\setlength{\tabcolsep}{2.5pt}
\begin{tabular}{llrrrrr}
\toprule
& & & \multicolumn{2}{c}{Effort RMS} & \multicolumn{2}{c}{Energy defect}\\
\cmidrule(lr){4-5}\cmidrule(lr){6-7}
Family & Updates & $n$ & Ref. & Gen. & Ref. & Gen.\\
\midrule
Stewart & 0 & 450 & 3.009 & 39.515 & 0.605 & 1.113 \\
Stewart & 50 & 450 & 3.009 & 40.513 & 0.605 & 1.640 \\
Delta & 0 & 359 & 0.733 & 12.295 & 0.209 & 0.613 \\
Delta & 50 & 424 & 0.733 & 12.319 & 0.203 & 1.206 \\
\bottomrule
\end{tabular}
\end{table}

\begin{table*}[t]
\caption{Spatial measurement controls on all 300 reference trajectories. Exact derivatives isolate the mechanics evaluator; 30 Hz configuration differences additionally test differentiation; reference RGB + CAD includes the frozen visual measurement path. Point RMSE is in mm, effort in N (Stewart) or N\,m (Delta), and energy in J. Pass counts use each family's 150 distinct trajectories.}
\label{app:spatial-controls}
\centering\small
\begin{tabular}{llrrrrr}
\toprule
Family & Input & Point & Effort & Energy & Effort pass & Energy pass\\
\midrule
Stewart & Exact states & 0.0000 & 0.0000 & 0.0064 & 150/150 & 150/150 \\
Stewart & 30 Hz differences & 0.0000 & 0.0477 & 0.0253 & 150/150 & 150/150 \\
Stewart & Reference RGB + CAD & 1.8287 & 3.0090 & 0.6052 & 149/150 & 48/150 \\
Delta & Exact states & 0.0000 & 0.0000 & 0.0056 & 150/150 & 150/150 \\
Delta & 30 Hz differences & 0.0000 & 0.0145 & 0.0253 & 150/150 & 150/150 \\
Delta & Reference RGB + CAD & 2.2475 & 0.7339 & 0.2020 & 91/150 & 63/150 \\
\bottomrule
\end{tabular}
\end{table*}

\setcounter{dbltopnumber}{3}
\renewcommand{\dbltopfraction}{0.9}
\makeatletter
\setlength{\@dblfptop}{0pt}
\setlength{\@dblfpsep}{12pt}
\makeatother

\section{Matched 64/256-pixel experiments}
\label{app:resolution}
\subsection{Shared observations and native training budgets}
The two resolutions sample the original $640\times640$ renders directly, using the same affine camera map, $1.30$ m field of view, frame times, actuator commands, physical labels, and assembly partitions. Each resolution contains 1,029 sequences: 813 training, 54 validation, 54 independent calibration, and 108 test sequences. Pixel-center spacing is $20.635/5.098$ mm at $64/256$ pixels. Localization can be subpixel, and the engineering tolerances remain identical across the two grids.

All four model--resolution combinations complete three seeds (42--44), giving twelve video fits. Each resolution is trained from scratch and evaluated at the fixed final epoch. Table~\ref{app:resolution-budget} records the architecture size and optimization budget. CON retains its native dense image layers; KeyCLD retains its encoder, renderer, potential, input map, and point masses. The comparison therefore follows native model scaling with resolution. The fifteen exact-state fits are shared controls because their physical trajectories and objectives are independent of image size.

\begin{table*}[t]
\caption{Native video training at both resolutions. Counts are per seed; all rows use seeds 42--44 and the same 813 training sequences. Resolution changes image-layer capacity, while epochs, batches, sequence exposures, and the model-specific dynamical objectives stay fixed.}
\label{app:resolution-budget}
\centering\small
\begin{tabular}{llrrrrrr}
\toprule
Model & Pixels & Parameters & Epochs & Batch & Updates & Sequence exposures & Forecast intervals\\
\midrule
KeyCLD & $64^2$ & 596,508 & 40 & 1 & 32,520 & 32,520 & 4 per window\\
KeyCLD & $256^2$ & 2,378,268 & 40 & 1 & 32,520 & 32,520 & 4 per window\\
CON & $64^2$ & 67,271,267 & 50 & 16 & 2,550 & 40,650 & 48 per sequence\\
CON & $256^2$ & 1,075,870,307 & 50 & 16 & 2,550 & 40,650 & 48 per sequence\\
\bottomrule
\end{tabular}
\end{table*}

KeyCLD's reconstruction loss sums over pixels. Multiplying it by $(64/N)^2$ maintains its weight relative to the four-step dynamics and heatmap losses as width $N$ changes; the latter two weights remain one. CON's RGB losses already average over pixels and retain all released weights. Full-sequence objectives are accumulated before each update, using framewise rematerialization to control memory. Both resolutions use FP32 model training and the same optimization schedules. Forecasting uses the three-frame prefix and the common physical-time grid.

\subsection{Visual, coordinate, and continuous-score comparisons}
Table~\ref{app:resolution-seeds} reports every seed and both measurement tracks. RGB and motion-mask errors average future frames 33--119, with the same temporal-variance mask rule at each resolution. Persistence is computed on the corresponding grid. Dividing each model's MSE by its same-resolution persistence MSE gives the normalized comparison in the main figure; it accounts for changes in the sampled image and foreground mask.

\begin{table*}[t]
\caption{All twelve native video fits, with 108 test trajectories per seed. RGB MSE is in $10^{-4}$ and motion-mask MSE in $10^{-3}$.}
\label{app:resolution-seeds}
\centering\small
\begin{tabular}{llrrrrr}
\toprule
Model & Pixels & Seed & RGB MSE & Motion-mask MSE & Native $S_{\rm comp}$ & Reconstructed $S_{\rm comp}$\\
\midrule
KeyCLD & 64 & 42 & 4.2396 & 5.5810 & 14.78 & 52.02 \\
KeyCLD & 64 & 43 & 5.5418 & 6.5761 & 21.14 & 54.32 \\
KeyCLD & 64 & 44 & 5.0970 & 6.0968 & 11.28 & 47.02 \\
KeyCLD & 256 & 42 & 4.8051 & 6.5578 & 33.37 & 54.32 \\
KeyCLD & 256 & 43 & 3.8003 & 4.9822 & 24.12 & 51.65 \\
KeyCLD & 256 & 44 & 4.9483 & 6.6122 & 9.41 & 53.33 \\
CON & 64 & 42 & 3.1102 & 4.5250 & 9.51 & 47.45 \\
CON & 64 & 43 & 3.1278 & 4.5120 & 3.53 & 46.95 \\
CON & 64 & 44 & 3.0027 & 4.5159 & 8.64 & 48.83 \\
CON & 256 & 42 & 3.1935 & 4.6833 & 9.29 & 49.74 \\
CON & 256 & 43 & 3.1488 & 4.6515 & 9.75 & 46.57 \\
CON & 256 & 44 & 3.1600 & 4.6289 & 13.84 & 48.47 \\
\bottomrule
\end{tabular}
\end{table*}

All 648 forecasts at 256 pixels are finite. Native physical coverage and joint qualification remain zero for both models at both resolutions. Table~\ref{app:resolution-components} explains the aggregate changes: velocity and acceleration consistency contribute most of the increased score, while physical point tracking remains poor. KeyCLD's native point RMSE changes from 229.2 to 233.9 mm, and CON's from 123.5 to 134.6 mm. Three-seed dispersion in Figure~\ref{app:resolution-ablation} complements the mean comparison.

Geometric reconstruction requires mean coordinate corrections of 120.49/128.13 mm for KeyCLD at 64/256 pixels and 73.18/79.17 mm for CON. It restores complete measurement coverage for 311/324 and 324/324 KeyCLD cases, respectively, and 308/324 and 311/324 CON cases. At 256 pixels, reconstructed torque residuals average 6.229/6.691 N\,m and energy defects 1.201/10.470 J for KeyCLD/CON. Joint qualification remains zero. These measurements connect the improved appearance scores to the continuing cost of obtaining physically admissible motion.

The six main-text time histories use S-A-109-P01 (low speed, seed 42), with the common crank readout and native derivatives at 240 Hz. Forecasting starts at 1.100 s; work is anchored at the last observed frame, 1.067 s.

\subsection{Independent 256-pixel reader and derivative ablations}
The independent U-Net retains widths 16/32/64/96, GroupNorm, and unconstrained soft-argmax joint coordinates. It is fitted on the 813 training sequences for 32 epochs with effective batch 128, Gaussian heatmap width 0.75 pixels, coordinate-loss weight 25, and learning rate $0.002$ decaying to $0.0002$. Training uses mixed precision with FP32 heatmaps/readouts. Evaluation uses FP32 with TF32 disabled. The frozen reader processes correct and generated RGB with the same settings.

On the same 108 test clips, unprojected Euclidean point RMSE on correct RGB falls from 2.127 mm at 64 pixels to 0.334 mm at 256 pixels. Both values average each trajectory's Euclidean RMS over the same 87 future frames (1.100--3.967 s), using the reader calibrated for that resolution. The corresponding 256-pixel forecast errors are 194.994/208.454 mm for KeyCLD/CON, averaged over three seed means. This coordinate comparison precedes reconstruction and differentiation; Table~\ref{app:resolution-reader} reports maximum-channel SI errors on the common interior window.

We compare Cartesian quintic differentiation, geometric reconstruction with Cartesian quintic derivatives, and geometric reconstruction with local cubic phase derivatives. Processing is selected on 54 validation sequences by position/velocity/acceleration error, subject to all 54 exact-coordinate references passing every SI criterion. The selected seven-frame cubic phase estimator is then checked on the independent 27 calibration and 27 verification sequences. It yields 16/27 and 14/27 joint passes on their correct RGB, while exact-coordinate replay retains 54/54 passes. Both resolutions use the same split roles; reader weights and derivative selection are frozen before testing.

The phase path improves correct-video $S_{\rm comp}$ from 43.70 (Cartesian) and 82.69 (projected Cartesian) to 90.00, with 66/108 joint passes. Reference errors are 0.144 mm for position, 0.116 N\,m for torque, and 0.057 J for energy against a 0.05 J energy target. Under the identical path, KeyCLD and CON give point errors of 183.029 and 181.873 mm, torque residuals of 6.565 and 6.585 N\,m, and energy defects of 10.835 and 1.485 J, respectively. Table~\ref{app:metric-paths} records the three coordinate statistics used in the main text.

\begin{table*}[p]
\begingroup
\caption{Continuous criterion scores (0--100), with equal seed/family/condition weighting. G/V/A are geometric, velocity, and acceleration consistency; F/E are effort and energy. The arithmetic mean of the six component columns is native $S_{\rm comp}$.}
\label{app:resolution-components}
\centering\small
\begin{tabular}{llrrrrrrrr}
\toprule
Model & Pixels & G & V & A & F & E & Point & Native & Reconstructed\\
\midrule
KeyCLD & 64 & 0.00 & 30.59 & 63.61 & 0.00 & 0.00 & 0.20 & 15.73 & 51.12 \\
KeyCLD & 256 & 0.00 & 56.81 & 76.79 & 0.00 & 0.00 & 0.21 & 22.30 & 53.10 \\
CON & 64 & 0.01 & 7.10 & 35.19 & 0.00 & 0.00 & 1.05 & 7.23 & 47.75 \\
CON & 256 & 0.01 & 7.22 & 57.60 & 0.00 & 0.00 & 0.91 & 10.96 & 48.26 \\
\bottomrule
\end{tabular}
\par
\endgroup
\par\vspace{12pt}
\begingroup
\captionof{table}{256-pixel measurement-path ablation on fixed videos. Correct RGB uses 108 unique test trajectories; each model uses 324 forecasts over three seeds. The same interior 80 frames (1.167--3.800 s) are evaluated throughout. Point is maximum-channel RMSE in mm; torque is in N\,m; energy is in J. Available-case means are balanced. Raw Cartesian force/energy means use only its 56 measurable reference clips; projected tracks use all registered clips.}
\label{app:resolution-reader}
\centering\small
\setlength{\tabcolsep}{4pt}
\begin{tabular}{llrrrrrr}
\toprule
Video & Measurement path & $S_{\rm comp}$ & Coverage & Joint & Point & Torque & Energy\\
\midrule
Correct RGB & Cartesian & 43.70 & 56/108 & 0/108 & 0.315 & 0.241 & 0.067 \\
Correct RGB & Projected Cartesian & 82.69 & 108/108 & 37/108 & 0.210 & 0.277 & 0.099 \\
Correct RGB & Projected phase & 90.00 & 108/108 & 66/108 & 0.144 & 0.116 & 0.057 \\
KeyCLD RGB & Cartesian & 0.65 & 0/324 & 0/324 & 185.503 & N/A & N/A \\
KeyCLD RGB & Projected Cartesian & 51.23 & 324/324 & 0/324 & 183.007 & 7.084 & 12.854 \\
KeyCLD RGB & Projected phase & 50.85 & 324/324 & 0/324 & 183.029 & 6.565 & 10.835 \\
CON RGB & Cartesian & 0.27 & 0/324 & 0/324 & 212.765 & N/A & N/A \\
CON RGB & Projected Cartesian & 52.77 & 324/324 & 0/324 & 181.825 & 6.520 & 1.356 \\
CON RGB & Projected phase & 50.99 & 324/324 & 0/324 & 181.873 & 6.585 & 1.485 \\
\bottomrule
\end{tabular}
\par
\endgroup
\par\vspace{12pt}
\centering
\includegraphics[width=.88\textwidth]{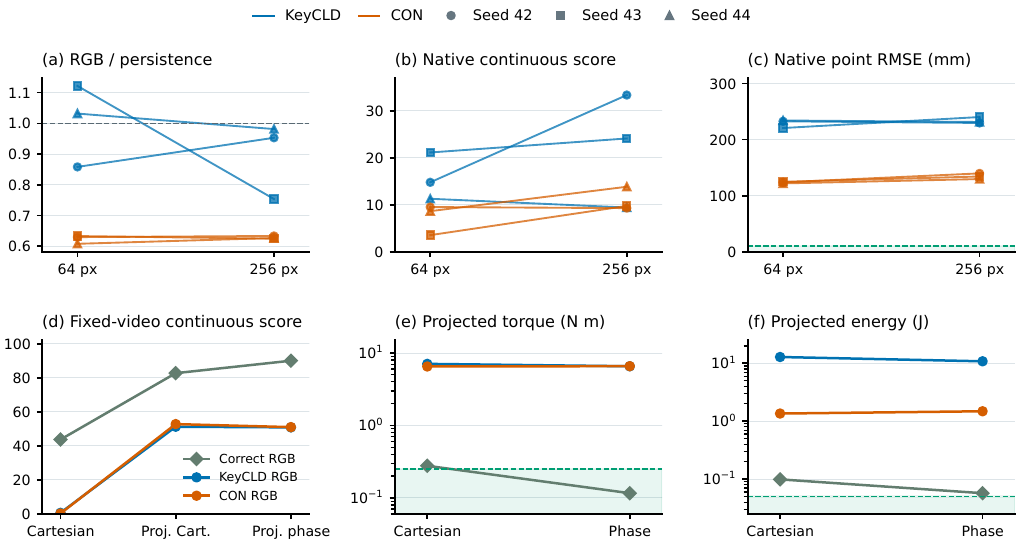}
\captionof{figure}{Resolution and measurement-path ablations. (a--c) Each line connects one seed's 64/256-pixel results within a model; RGB uses same-resolution persistence, and point is native maximum-channel RMSE. (d--f) Frozen 256-pixel videos pass through Cartesian, projected-Cartesian, and projected-phase measurements; the score panel compares all three choices, while torque/energy compare the two projected paths on identical support. Projected reference coverage is 108/108. Shaded bands mark engineering tolerances.}
\label{app:resolution-ablation}

\end{table*}

\begin{table*}[t]
\caption{Planar 256-pixel point-error definitions. All entries average trajectory RMSEs, with model results balanced across seeds, families, and conditions. The native-grid reconstruction uses each model's declared physical readout; the other paths use the independent frozen RGB reader.}
\label{app:metric-paths}
\centering\small
\setlength{\tabcolsep}{4pt}
\begin{tabular}{llllrr}
\toprule
Measurement path & Future support & Coordinate statistic & Reference (mm) & KeyCLD (mm) & CON (mm)\\
\midrule
Native-grid reconstruction & 689 samples, 240 Hz & Maximum-channel RMS & -- & 198.19 & 167.35\\
Unprojected RGB reader & 87 frames, 1.100--3.967 s & Euclidean RMS & 0.334 & 194.994 & 208.454\\
Projected-phase RGB reader & 80 frames, 1.167--3.800 s & Maximum-channel RMS & 0.144 & 183.029 & 181.873\\
\bottomrule
\end{tabular}
\end{table*}

\subsection{Target sensitivity on saved residuals}
\label{app:target-sensitivity}
Table~\ref{app:thresholds} multiplies all six engineering targets by $0.5$, $1$, or $2$ and recomputes continuous scores and joint passes from saved per-trajectory residuals. Readouts, reconstruction, sample support, availability masks, and case counts stay fixed. Scores follow the weighting in Section~\ref{app:measurement}; the reference RGB row uses 108 distinct test clips at seed 42.

RK4's score advantage over Euler is 6.32, 11.65, and 17.90 points at the three scales. The listed model trajectories have zero joint passes throughout. Reference RGB joint passes rise from 34/108 to 66/108 and 86/108, showing the dependence of target attainment on the engineering scale. Native and reconstructed component scores and pass rates for each scale accompany the analysis tables.

\begin{table*}[t]
\caption{Target sensitivity on fixed measurements. Coverage is complete six-criterion availability and is constant across scales. Joint entries count cases passing every scaled target. CAD and reconstructed rows apply geometric interventions to saved forecasts.}
\label{app:thresholds}
\centering\small
\setlength{\tabcolsep}{4pt}
\begin{tabular}{llrrrrrrr}
\toprule
& & & \multicolumn{3}{c}{$S_{\rm comp}$} & \multicolumn{3}{c}{Joint passes}\\
\cmidrule(lr){4-6}\cmidrule(l){7-9}
Model / reference & Track & Coverage & $0.5\times$ & $1\times$ & $2\times$ & $0.5\times$ & $1\times$ & $2\times$\\
\midrule
KeyCLD Euler & State native & 324/324 & 51.35 & 52.65 & 54.28 & 0 & 0 & 0 \\
KeyCLD RK4 & State native & 324/324 & 57.67 & 64.30 & 72.19 & 0 & 0 & 0 \\
KeyCLD 64 px & Native & 0/324 & 10.68 & 15.73 & 21.56 & 0 & 0 & 0 \\
KeyCLD 64 px & Reconstructed & 311/324 & 49.79 & 51.12 & 52.96 & 0 & 0 & 0 \\
KeyCLD 256 px & Native & 0/324 & 18.85 & 22.30 & 26.21 & 0 & 0 & 0 \\
KeyCLD 256 px & Reconstructed & 324/324 & 51.67 & 53.10 & 55.37 & 0 & 0 & 0 \\
CON 64 px & Native & 0/324 & 3.73 & 7.23 & 12.66 & 0 & 0 & 0 \\
CON 64 px & Reconstructed & 308/324 & 47.59 & 47.75 & 48.33 & 0 & 0 & 0 \\
CON 256 px & Native & 0/324 & 6.29 & 10.96 & 16.56 & 0 & 0 & 0 \\
CON 256 px & Reconstructed & 311/324 & 48.08 & 48.26 & 48.87 & 0 & 0 & 0 \\
V-JEPA + AC Stewart & Native & 0/450 & 1.72 & 6.20 & 18.24 & 0 & 0 & 0 \\
V-JEPA + AC Stewart & CAD & 450/450 & 50.51 & 55.11 & 63.47 & 0 & 0 & 0 \\
V-JEPA + AC Delta & Native & 0/450 & 3.55 & 10.40 & 23.83 & 0 & 0 & 0 \\
V-JEPA + AC Delta & CAD & 432/450 & 44.78 & 50.05 & 56.45 & 0 & 0 & 0 \\
LingBot 0 Stewart & Native & 0/450 & 1.56 & 5.67 & 17.03 & 0 & 0 & 0 \\
LingBot 0 Stewart & CAD & 450/450 & 49.37 & 54.60 & 64.14 & 0 & 0 & 0 \\
LingBot 0 Delta & Native & 0/450 & 0.78 & 2.80 & 8.10 & 0 & 0 & 0 \\
LingBot 0 Delta & CAD & 359/450 & 37.25 & 43.37 & 51.41 & 0 & 0 & 0 \\
LingBot 50 Stewart & Native & 0/450 & 1.66 & 6.01 & 17.76 & 0 & 0 & 0 \\
LingBot 50 Stewart & CAD & 450/450 & 50.15 & 54.31 & 62.87 & 0 & 0 & 0 \\
LingBot 50 Delta & Native & 0/450 & 0.94 & 3.39 & 9.88 & 0 & 0 & 0 \\
LingBot 50 Delta & CAD & 424/450 & 46.41 & 51.92 & 60.19 & 0 & 0 & 0 \\
Reference RGB 256 px & Projected phase & 108/108 & 83.10 & 90.00 & 95.07 & 34 & 66 & 86 \\
\bottomrule
\end{tabular}

\end{table*}
\FloatBarrier

\end{document}